\documentclass[]{article}

\usepackage[utf8]{inputenc}
\usepackage[english]{babel}

\usepackage{hyperref}
\hypersetup{
	colorlinks=true,
	linkcolor=blue,
	filecolor=magenta,      
	urlcolor=cyan,
}

\usepackage{graphicx}
\usepackage{subfigure}
\usepackage{enumitem}

\usepackage{pifont}

\usepackage{booktabs}

\usepackage{makecell}

\newcommand*\rot{\rotatebox{90}}

\usepackage{dirtree}

\newcolumntype{H}{>{\setbox0=\hbox\bgroup}c<{\egroup}@{}}

\usepackage{amsmath}

\usepackage[letterspace=-35]{microtype}

\title{Cityscapes-Panoptic-Parts \\ and PASCAL-Panoptic-Parts \\ datasets for Scene Understanding}
\author{Panagiotis Meletis\footnote{\lsstyle{Panagiotis Meletis (\texttt{p.c.meletis@tue.nl}) and Xiaoxiao Wen (\texttt{xiaoxiao.wen@student.uva.nl}) can be contacted for any suggestions or questions. Authors contributed equally.}}~\footnotemark[2]~, Xiaoxiao Wen\footnotemark[1]~\footnotemark[3]~, \\ Chenyang Lu\footnotemark[2]~, Daan de Geus\footnotemark[2]~, Gijs Dubbelman\footnotemark[2]}
\date{\footnotemark[2]~ Eindhoven University of Technology \\ \footnotemark[3]~ University of Amsterdam}

\begin{document}

\maketitle

\begin{abstract}
In this technical report, we present two novel datasets for image scene understanding. Both datasets have annotations compatible with panoptic segmentation and additionally they have part-level labels for selected semantic classes. This report describes the format of the two datasets, the annotation protocols, the merging strategies, and presents the datasets statistics. The datasets labels together with code for processing and visualization will be published at \href{https://github.com/tue-mps/panoptic_parts}{github.com/tue-mps/panoptic\_parts}.
\end{abstract}

\section{Introduction}

\begin{figure}[t]
	\centering
	\includegraphics[width=0.8\linewidth]{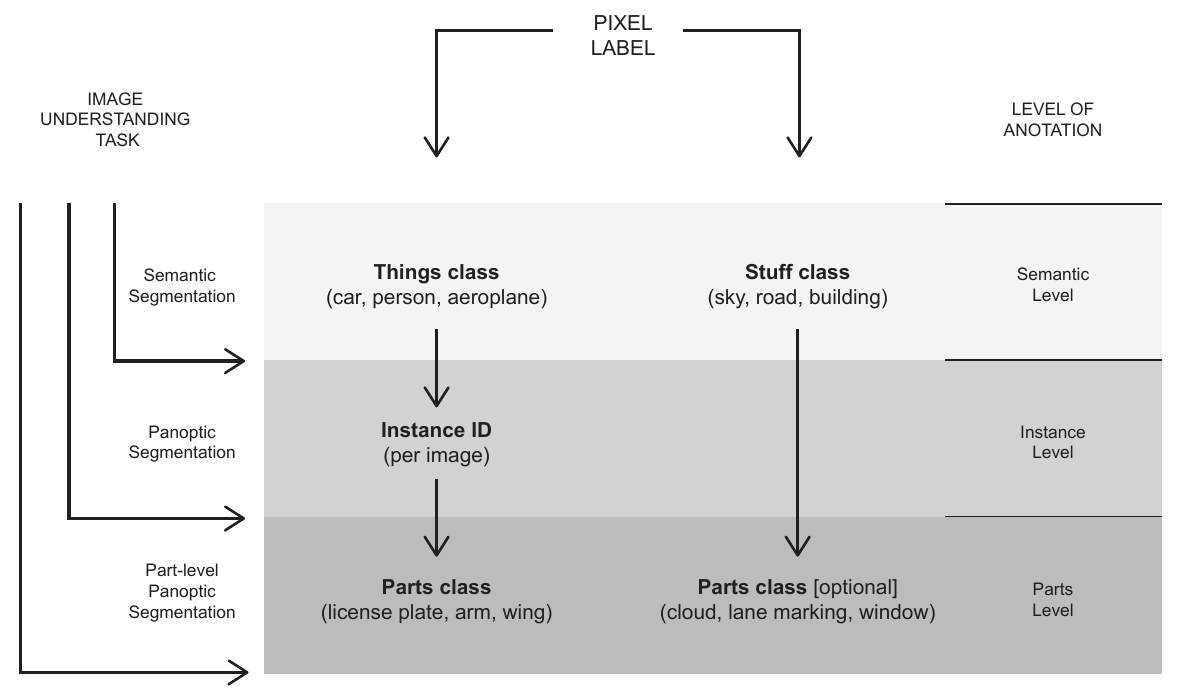}
	\vspace{-5px}
	\caption{The label hierarchy according to which we provide annotations for and the related image understanding tasks that can be used.}
	\label{fig:tasks}
\end{figure}

Our goal is to provide two datasets with consistent, multi-level annotations covering a wide spectrum of image understanding tasks, both holistic (panoptic segmentation), as well as detailed ones (semantic segmentation, part-level segmentation, object detection).

To create the datasets we use existing published datasets, and we either augment them with our manual annotations or merged them, to create a multi-level hierarchy of labels (see Figure~\ref{fig:tasks}). The whole hierarchy is given on the same set of images, and is free of any conflicts or inconsistencies that would arise if the constituent datasets were used separately. This enables researchers to use our datasets in a wide range of image understanding problems, and create and test their algorithms on an arbitrary level of label granularity.

Each pixel in the datasets is labeled on three levels: semantic level, instance level, and parts level. On the semantic level we maintain the intuitive separation that panoptic segmentation introduces between \textit{stuff} and \textit{things} classes~\cite{Kirillov2019PS}.
Currently, we provide part-level annotations only for the \textit{things} classes, however we note that, on a conceptual level, \textit{stuff} classes may also have parts and we leave this for future research.

The first dataset, \textit{Cityscapes-Panoptic-Parts} extends the established Cityscapes dataset for urban street scenes understanding, with our manual annotations for 23 part-level classes. The second dataset, \textit{PASCAL-Panoptic-Parts} extends the recognized PASCAL dataset for generic, everyday life scenes by merging two datasets, PASCAL-Parts, which has part-level annotations only for the classes of PASCAL and PASCAL-Context with semantic annotations only. A comparison with related datasets can be seen in Table~\ref{tab:related-datasets}.



\begin{table}[h]
    \footnotesize
	\setlength\tabcolsep{4.0pt}
	\label{tab:related-datasets}
	\caption{Dataset statistics for related parts segmentation datasets and our proposed dataset. The difference of the average number of instances with parts between Cityscapes and \textit{Cityscapes-Panoptic-Parts} is due to small instances and parts that are not distinguishable.
	\textit{PASCAL-Context} has 459 semantic classes but only 59 of them are included in the official split.}
	\centering
	\begin{tabular}{lcccccHHccc}
		\toprule
		\textbf{Dataset} & \rot{\makecell[c]{\textbf{Inst. aware}}} & \rot{\makecell[c]{\textbf{Panop. aware}}} & \rot{\makecell[l]{\textbf{Stuff classes}}} & \rot{\makecell[l]{\textbf{Things classes}}} & \rot{\makecell[l]{\textbf{Parts classes}}} & \rot{\makecell[l]{\textbf{Human parts}}} & \rot{\makecell[l]{\textbf{Vehicle parts}}} & \rot{\makecell[l]{\textbf{\# Images} \\ \textbf{train / val}}} & \rot{\makecell[l]{\textbf{Average} \\ \textbf{image size}}} & \rot{\makecell[l]{\textbf{Average} \\ \textbf{\# insts.}}} \\
		\midrule
		PASCAL-Context~\cite{mottaghi14context} & - & - & 59 & - & - & - & - & 4998/5105 & 387 x 470 & - \\
		LIP~\cite{liang2018look} & - & - & - & 1 & 20 & 20 & - & 30.5k/10k & 325 x 240 & 1 \\
		CIHP~\cite{gong2018instance} & \ding{51} & - & - & 1 & 20 & 20 & - & 28.3k/5k & 484 x 578 & 3.4 \\
		MHP v2.0~\cite{zhao2018understanding} & \ding{51} & - & - & 1 & 59 & 59 & - & 15.4k/5k & 644 x 718 & 3 \\
		PASCAL-Person-Parts~\cite{chen2014detect} & \ding{51} & - & - & 1 & 6 & 6 & - & 1716/1817 & 387 x 470 & 2.2 \\
		PASCAL-Parts~\cite{chen2014detect} & \ding{51} & - & - & 20 & 193 & 24 & 57 & 4998/5105 & 387 x 470 & 4 \\
		Cityscapes~\cite{Cordts2016Cityscapes} & \ding{51} & \ding{51} & 23 & 8 & - & - & - & 2975/500 & 1024 x 2048 & \textbf{16.2} \\
		\midrule
		\textbf{This work} & & & & & & & & & & \\
		\textbf{PASCAL-Panoptic-Parts} & \ding{51} & \ding{51} & 100 & 20 & 193 & 24 & 57 & 4998/5105 & 387 x 470 & 4 \\
		\textbf{Cityscapes-Panoptic-Parts} & \ding{51} & \ding{51} & 23 & 8 & 23 & 4 & 5 & 2975/500 & 1024 x 2048 & \textbf{15.6} \\
		\bottomrule
	\end{tabular}
\end{table}

\section{Cityscapes-Panoptic-Parts Dataset}
\label{sec:dataset}

Cityscapes dataset~\cite{Cordts2016Cityscapes} is a large-scale dataset of real-world urban scenes recorded in Germany and neighboring countries. Its densely annotated portion consists of three per-pixel manually labeled splits, i.e., training, validation, and test, which contain 2975, 500, and 1525 images, respectively. The complexity of urban scenes and the variety in number and pose of traffic participants, together with the fact that it is thoroughly studied, make it a perfect candidate for use in image and object understanding research.

\begin{figure}[h]
	\centering
	\includegraphics[width=0.48\linewidth]{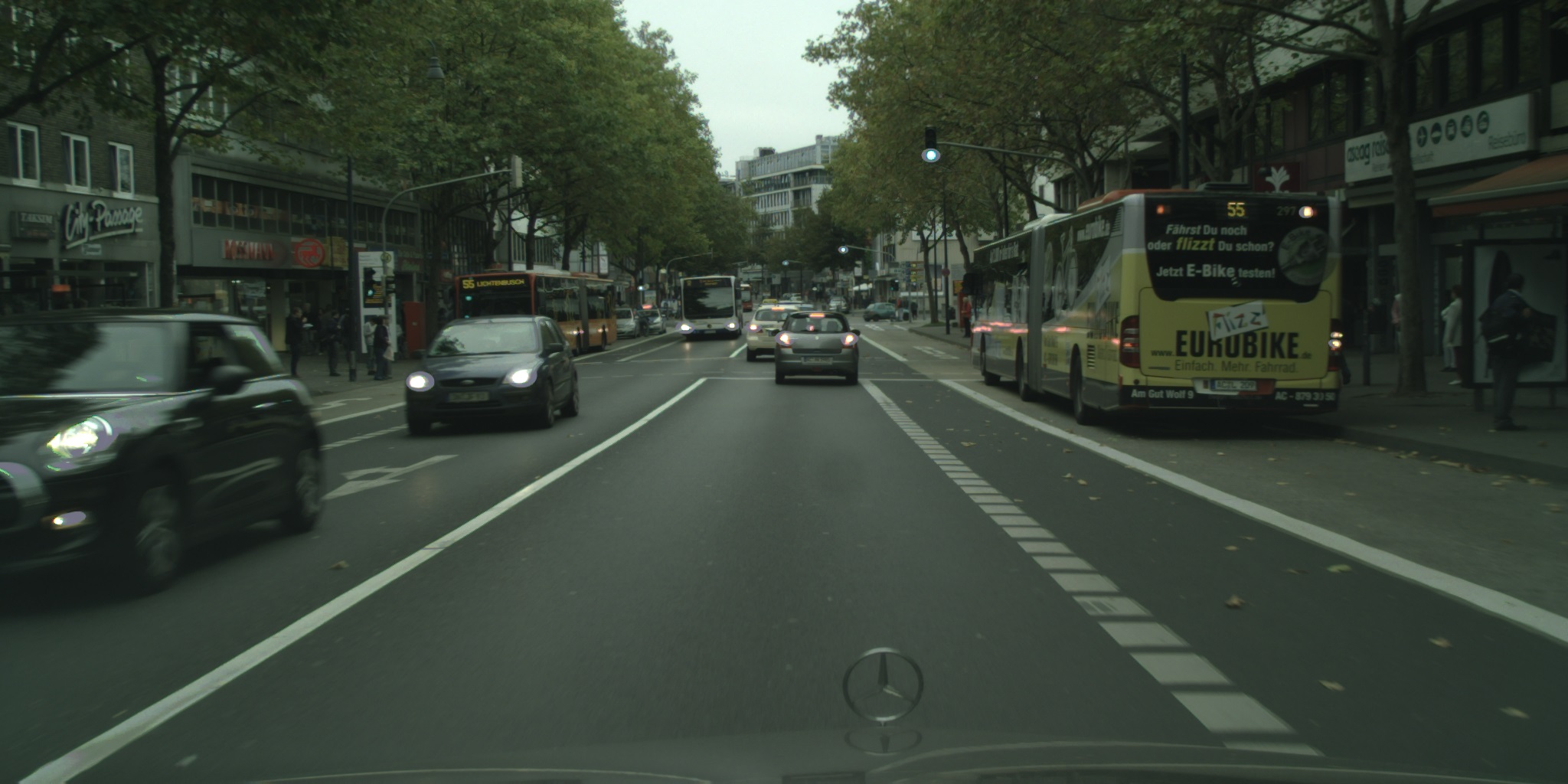}
	~\includegraphics[width=0.48\linewidth]{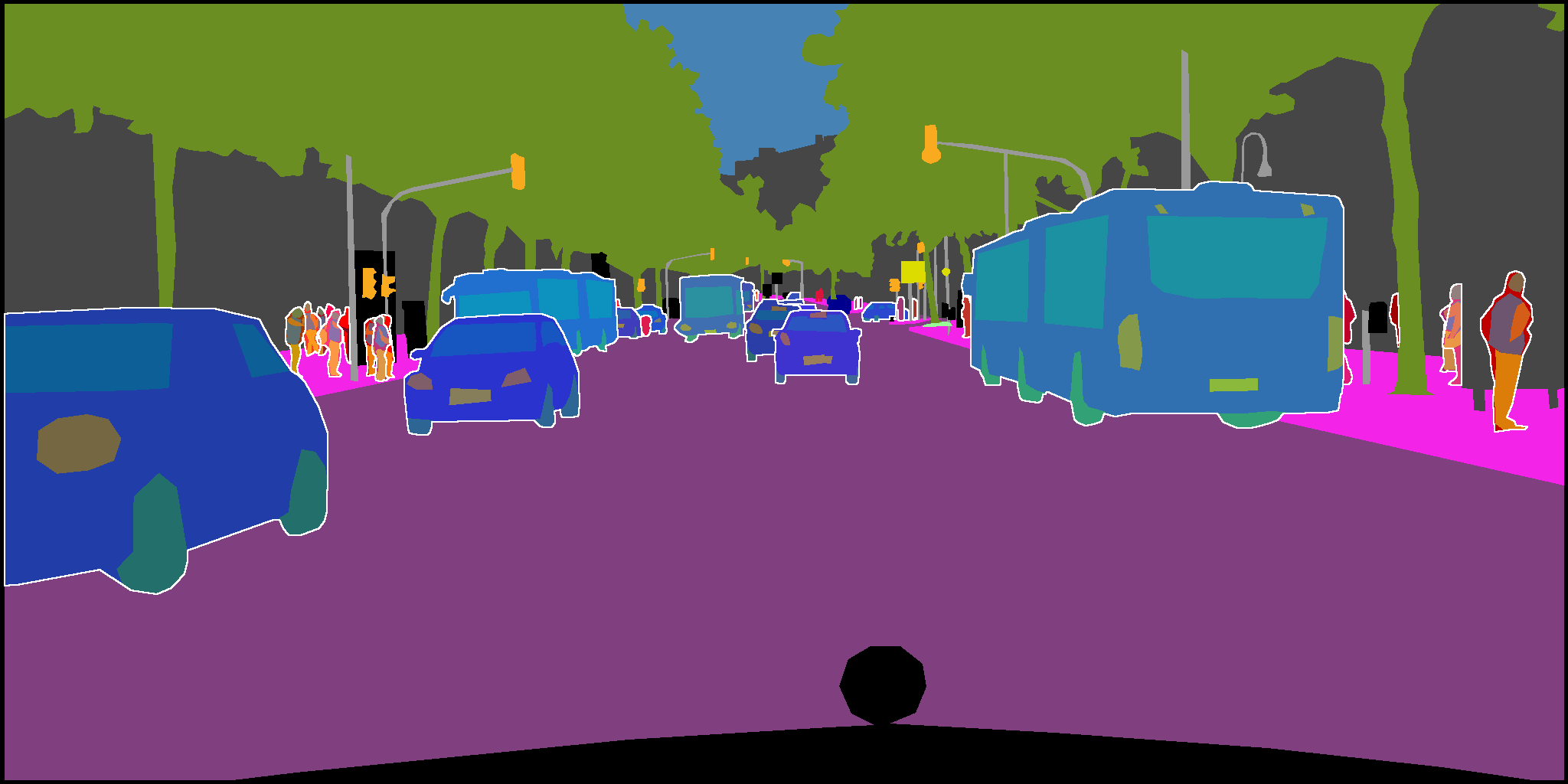}\\
	\includegraphics[width=0.48\linewidth]{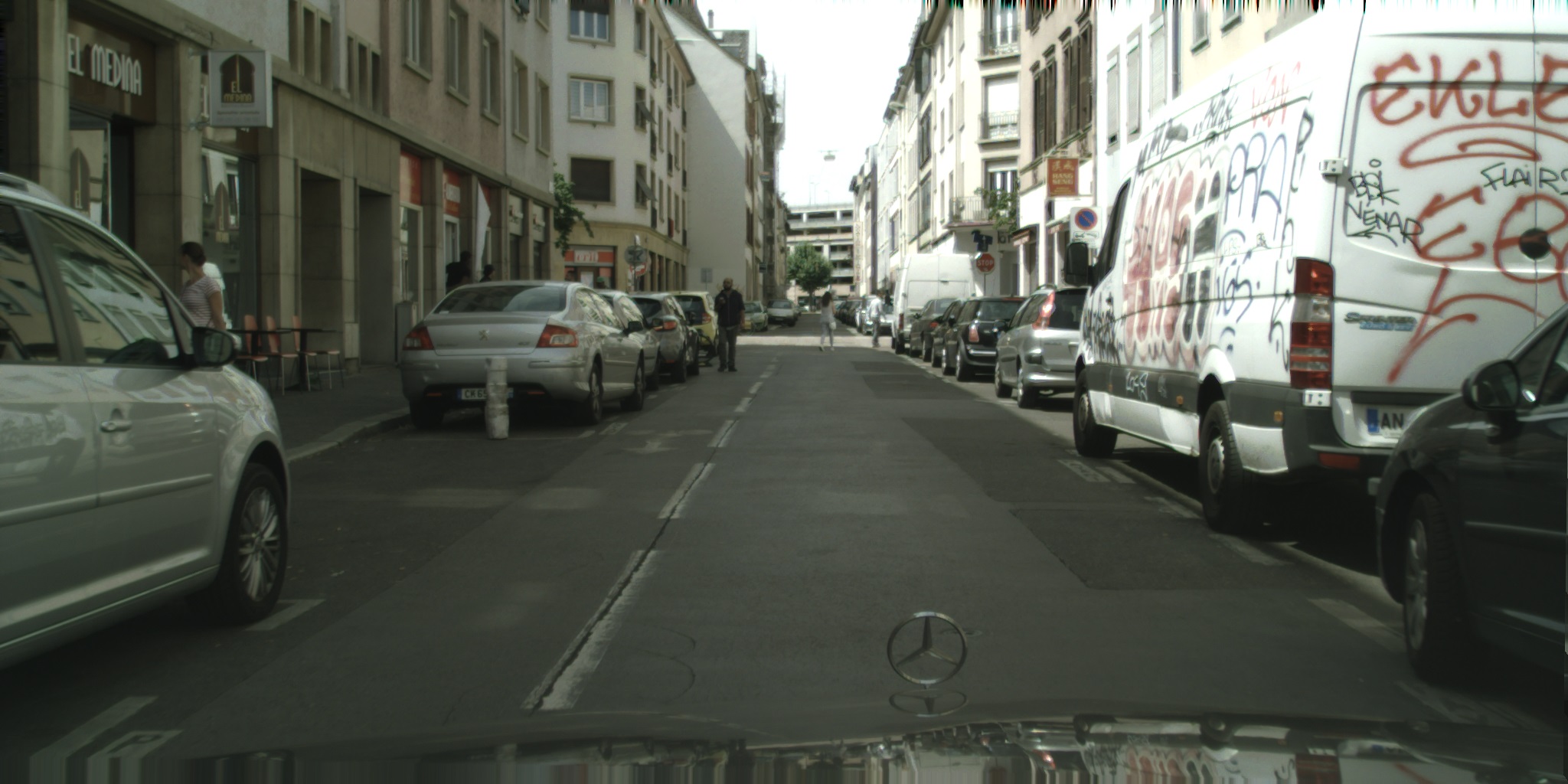}
	~\includegraphics[width=0.48\linewidth]{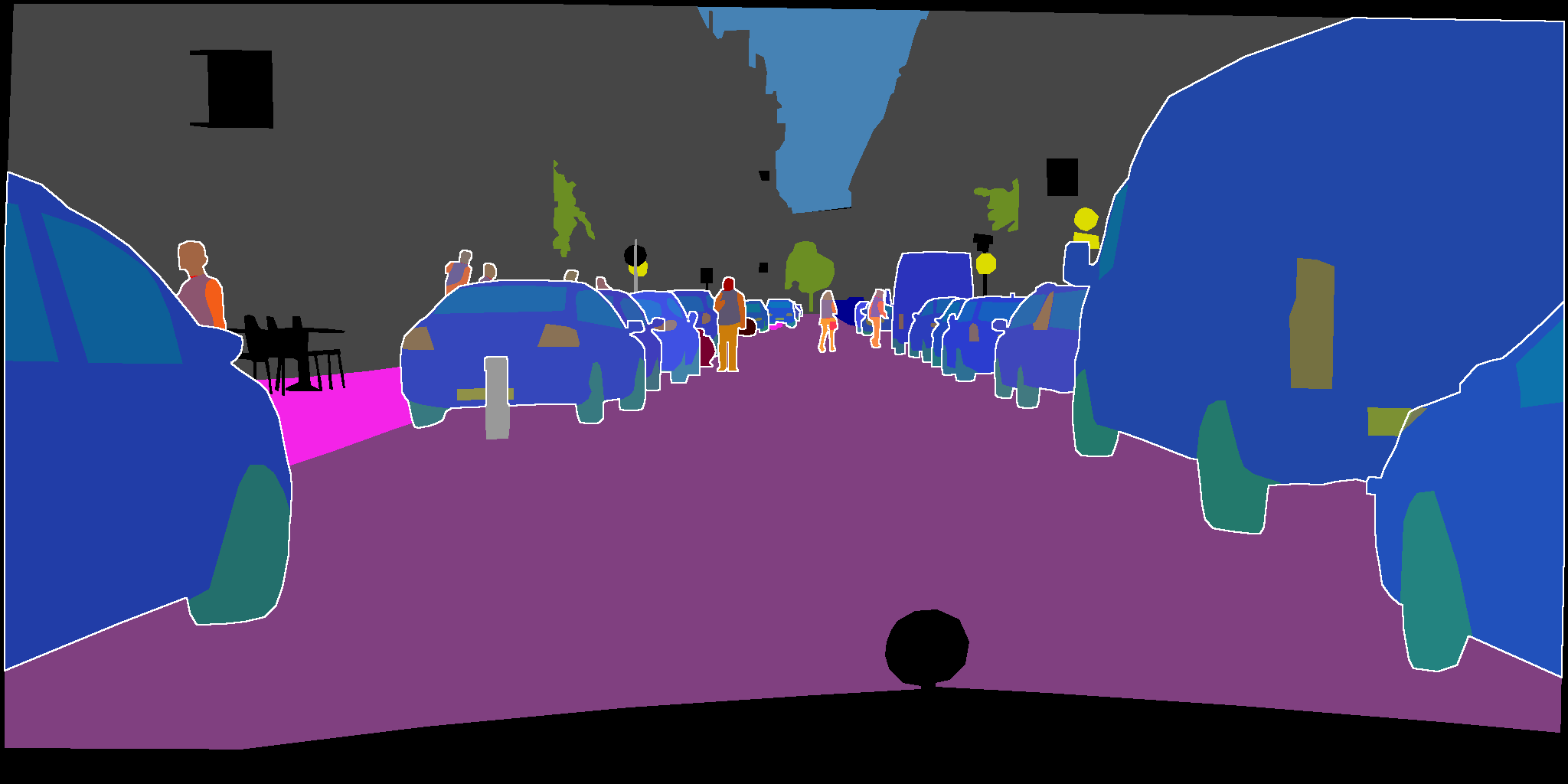}\\
	\includegraphics[width=0.48\linewidth]{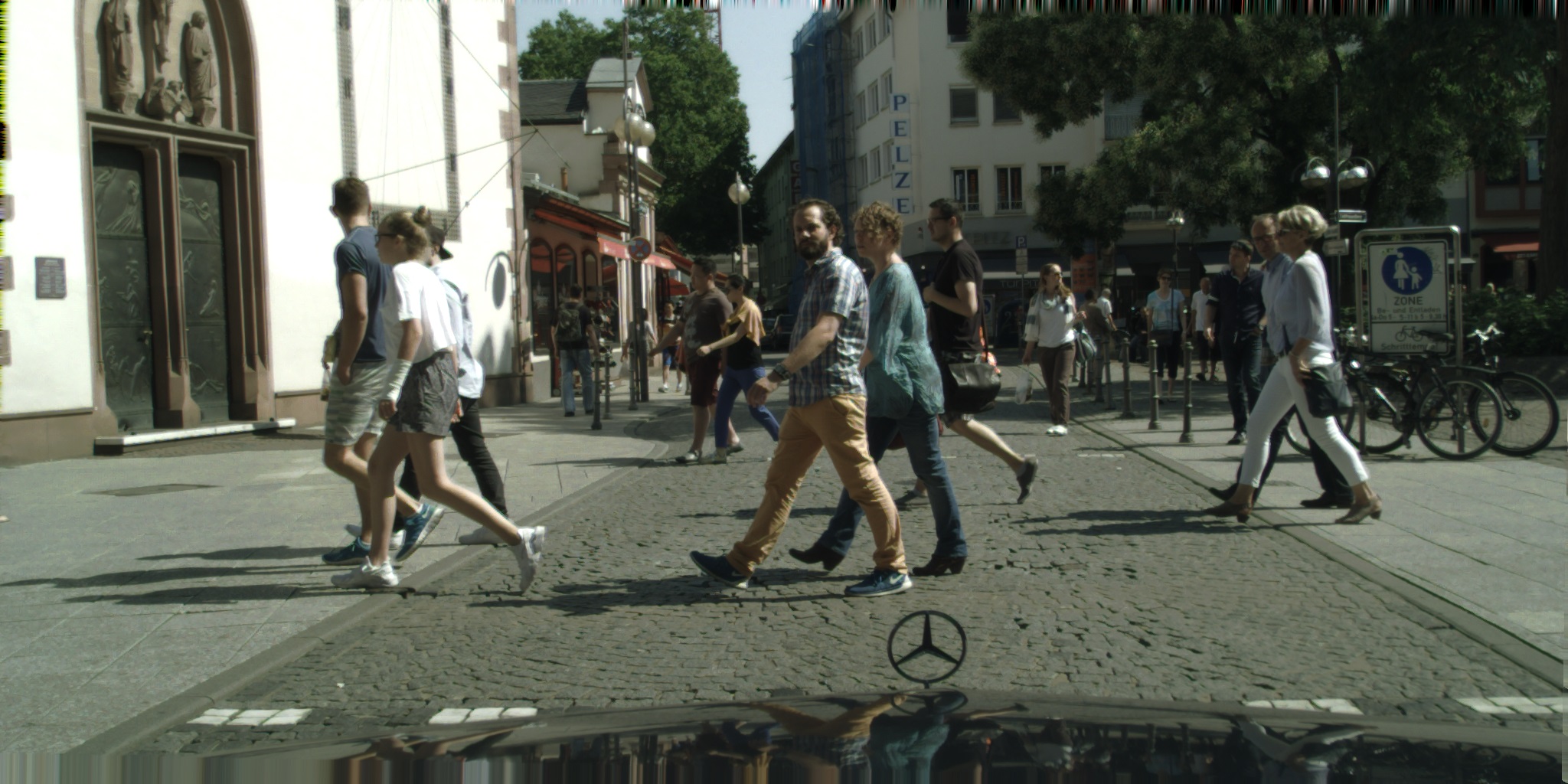}
	~\includegraphics[width=0.48\linewidth]{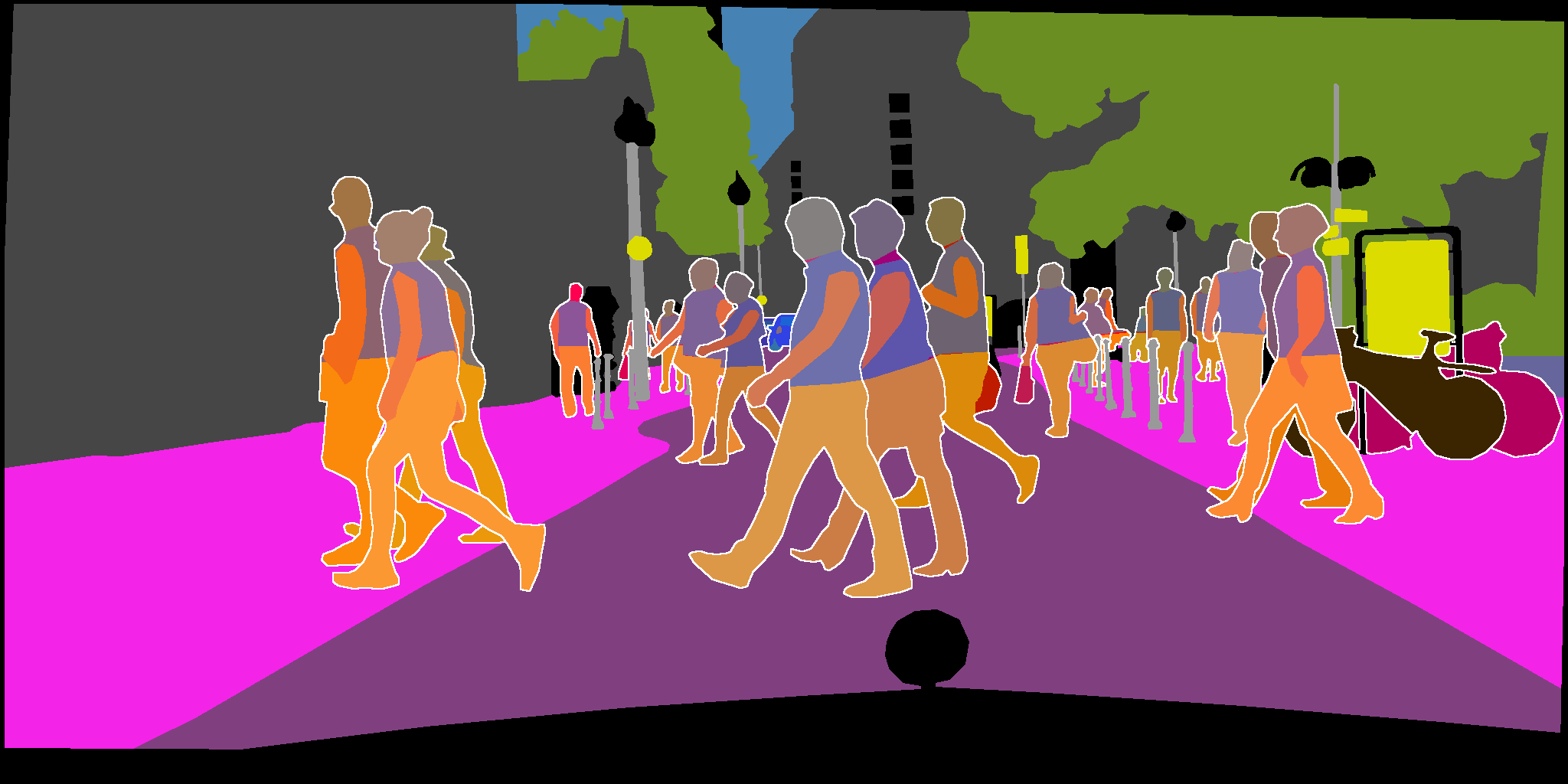}\\
	\includegraphics[width=0.48\linewidth]{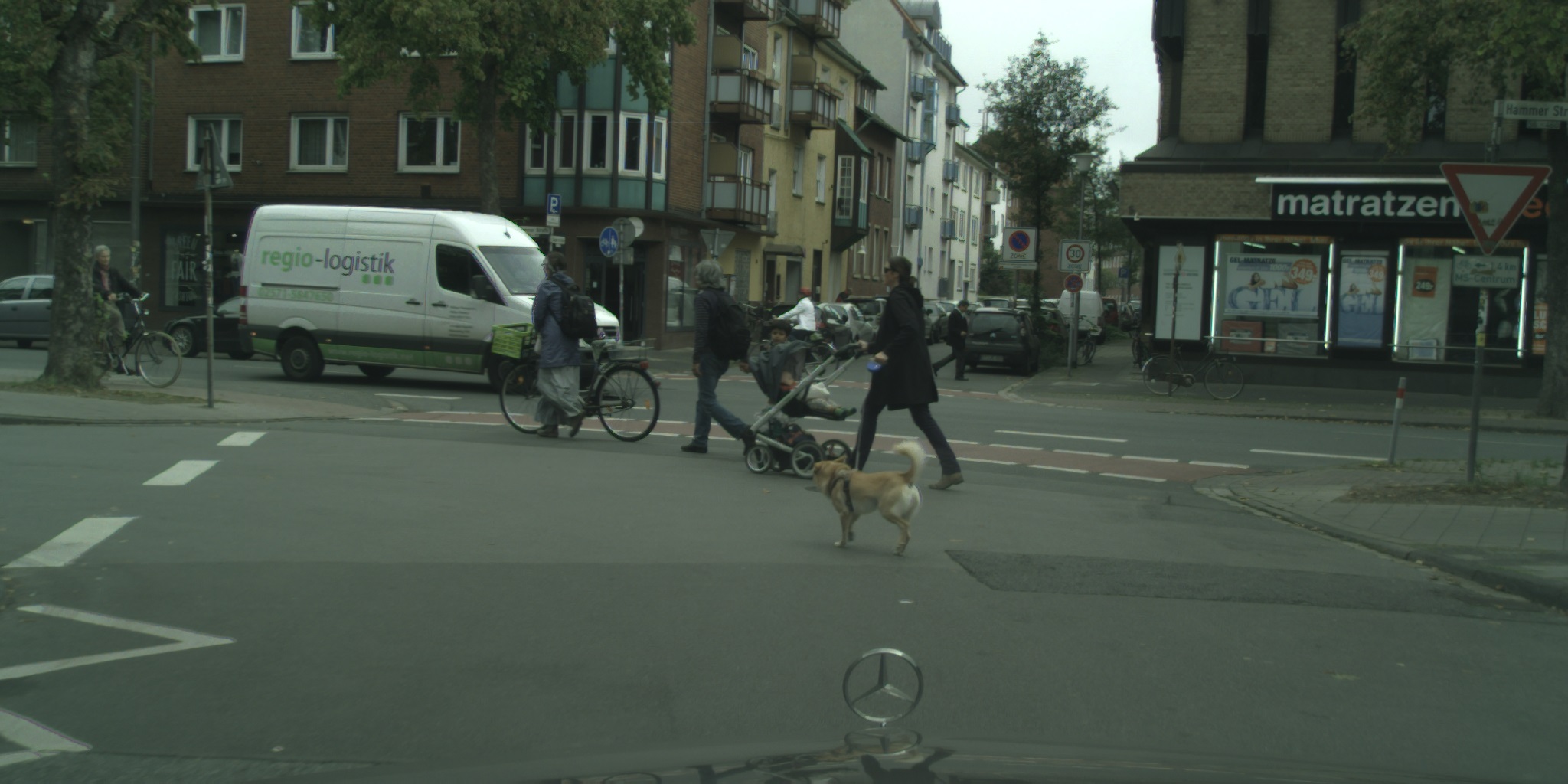}
	~\includegraphics[width=0.48\linewidth]{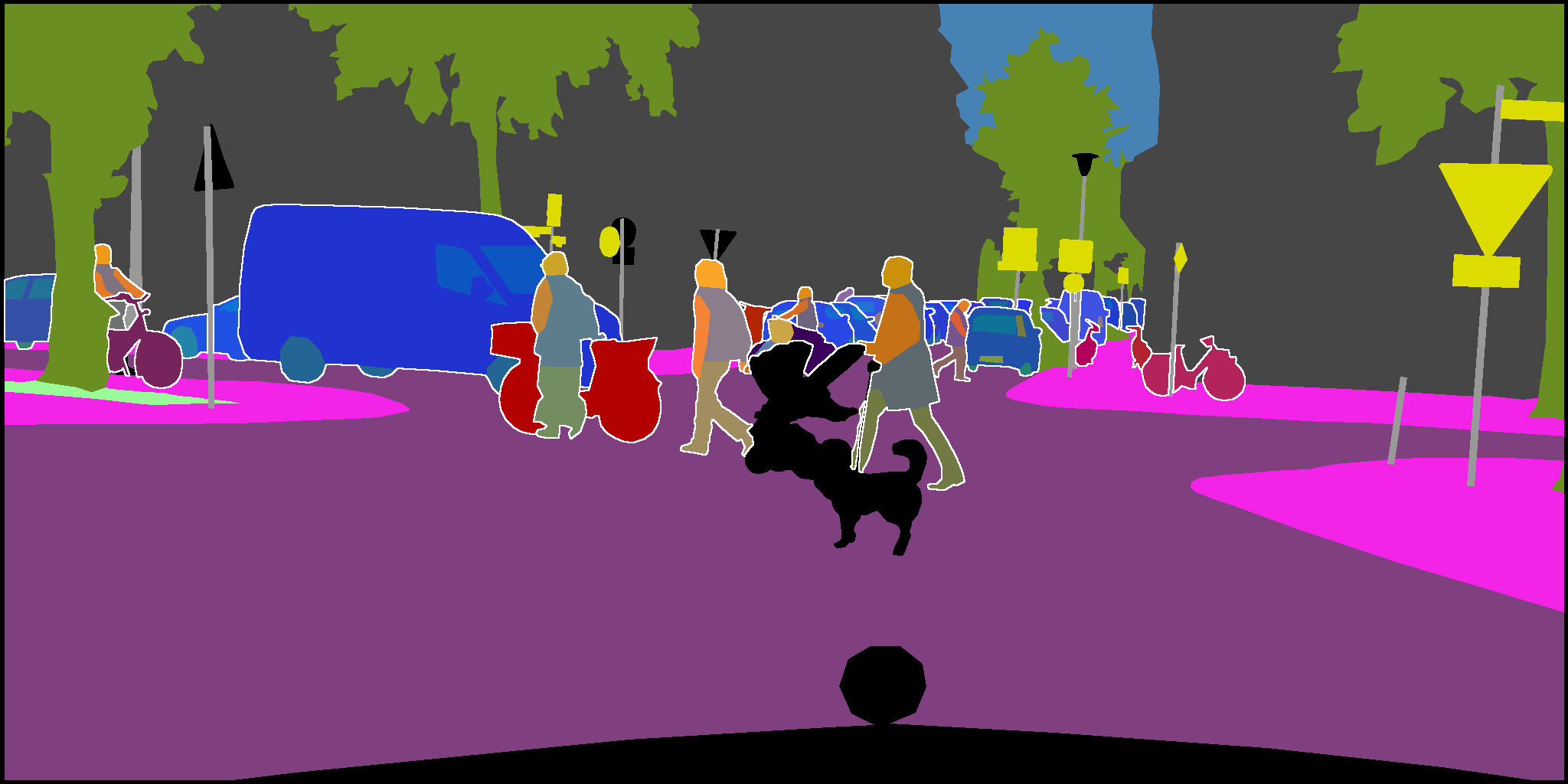}\\
	\vspace{-5pt}
	\caption{Examples from Cityscapes-Panoptic-Parts images and labels. Top two rows: training split. Bottom two rows: validation split.}
	\label{fig:gt_examples}
\end{figure}

In this work, we introduce \textit{Cityscapes-Panoptic-Parts}, a new dataset that extends Cityscapes with part-level annotations, in order to facilitate part-level object understanding in complex urban environments. Both Cityscapes and \textit{Cityscapes-Panoptic-Parts} have labels compatible with panoptic segmentation~\cite{Kirillov2019PS}. In Figure~\ref{fig:gt_examples}, we provide examples of images and labels from \textit{Cityscapes-Panoptic-Parts}.

Our main contributions with respect to the original Cityscapes dataset are: 1) a rich and detailed set of manual part-level annotations 2) a procedure for combining existing Cityscapes labels with our new part-level annotations, which does not create conflicts and minimizes the extra annotation time per image to an average of 7 minutes, described in Section~\ref{ssec:annot-proc}, and 3) a hierarchical encoding of the part-level panoptic labels explained in Section~\ref{ssec:data-format}.

\subsection{Dataset format}
\label{ssec:data-format}
We have manually annotated 5 semantic classes of Cityscapes with 9 parts classes in total. Specifically, the two human classes (person, rider) with 4 part-level classes, namely \textit{torso}, \textit{head}, \textit{arms}, \textit{legs}, and three vehicle classes (car, truck, bus) with 5 part-level classes, namely \textit{windows}, \textit{wheels}, \textit{lights}, \textit{license plate}, and \textit{chassis} are annotated. \textit{Cityscapes-Panoptic-Parts} is the first public dataset, to the best of our knowledge, with part-level panoptic labels.

We designed the labeling protocol and format with the following goals: 1) maximizing the \textit{compatibility} with existing datasets, 2) providing a \textit{compact} part-level panoptic label representation, which allows for future extensions, and 3) incorporating the \textit{hierarchical nature} of part-level panoptic segmentation into the label format. These requirements led to a labeling scheme that is compatible with the Cityscapes panoptic format and extends the single-file ground truth format in an intuitive hierarchical manner.

\begin{figure}[t]
	\centering
	\includegraphics[width=\linewidth, trim={2.0cm 0.1cm 0.2cm 0.2cm}, clip]{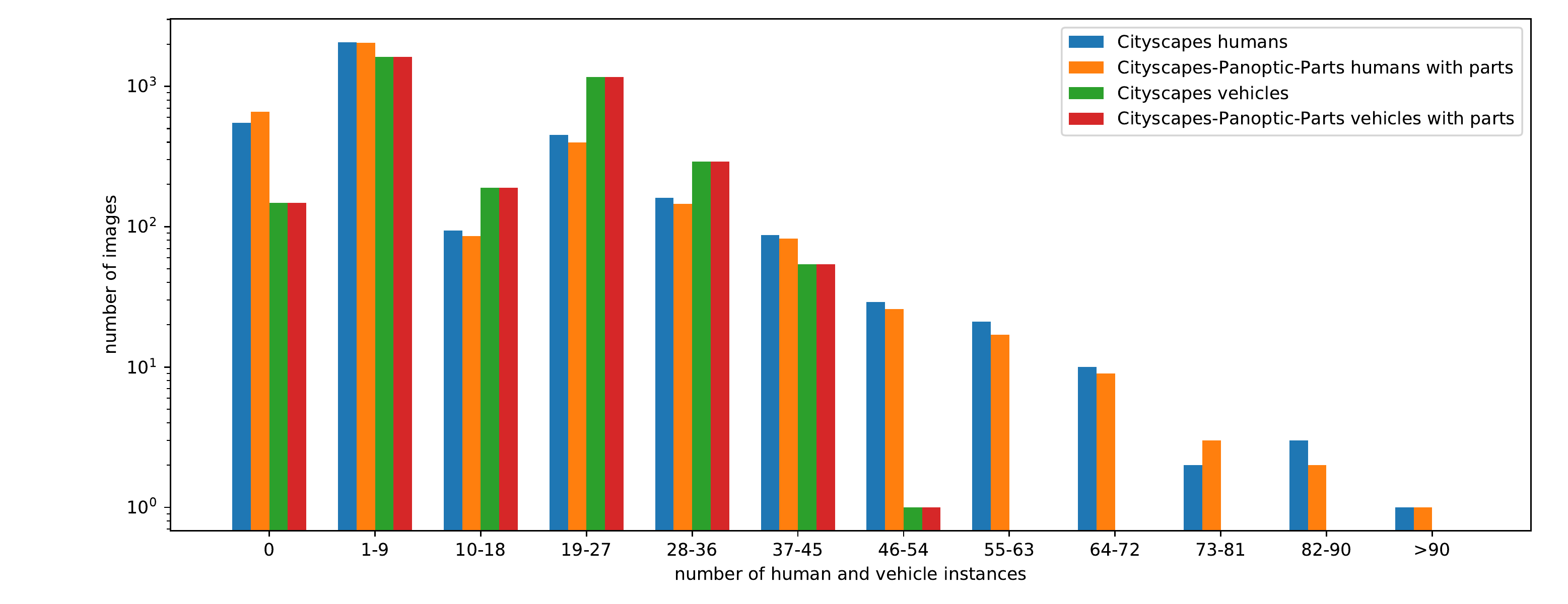}
	\vspace{-15pt}
	\caption{Dataset statistics visualized as a histogram for the original Cityscapes dataset and the new \textit{Cityscapes-Panoptic-Parts} split over images with humans and images with vehicles. Each bar shows the number of images in a dataset split that contains a certain number of road participants. The discrepancy between \textit{Cityscapes-Panoptic-Parts} humans and Cityscapes humans distributions is due to small instances that have small or indistinguishable parts, or contradictory labels and is explained in details in Section~\ref{ssec:annot-proc}.}
	\label{fig:num_images_hist}
\end{figure}

\subsection{Annotation procedure}
\label{ssec:annot-proc}
We employed a team of 11 annotators and three experts, and we split the load such that every Cityscapes city is labeled by multiple annotators to ensure that human bias for parts class definitions and error are minimized per-city. The procedure for creating the part-level panoptic ground truth has the following steps:
\begin{enumerate}
	\item An image is masked according to human and vehicle Cityscapes class labels, and these two masked images are provided to the annotators.
	\item The annotators create polygons to label part regions. The polygons at the boundaries of objects and outside of the masked pixels do not have to be precise since those areas can be discarded in post-processing.
	\item The annotations pass through automatic post-processing, which merges original Cityscapes annotations and our parts annotations into our hierarchical (three-level) format using the object masks when required.
\end{enumerate}
Using the procedure above, our parts annotations are guaranteed not to overlap with existing Cityscapes annotations that do not correspond to objects. Moreover, for regions or instances for which it is difficult to provide part-level annotations due to their small size, indistinguishable parts, or contradictory labels the original (two-level) panoptic labels are maintained. For example, although some backpack pixels in the original Cityscapes are labeled with the person class, we do not provide part-level annotations for them and maintain the original labels. All the aforementioned reasons introduce a discrepancy in the distributions for the human class in Figure~\ref{fig:num_images_hist}.

\subsection{Dataset statistics}
We have annotated 2975 images from the training split and 500 images from the validation split. Since the test split is not public, we cannot provide annotations. In addition to the statistics provided in the main paper, we also investigate scene complexity. We compute the frequency of images with a certain number of traffic participants, of the human or vehicle categories in the histogram of Figure~\ref{fig:num_images_hist}. All vehicle instances of \textit{Cityscapes-Panoptic-Parts} are annotated with at least one parts class, while for human instances, only a few are not labeled at the part-level.

\begin{figure}[t]
	\begin{center}
		\includegraphics[width=0.65\linewidth, trim={0.0cm 0.1cm 0.0cm 0.0cm}, clip]{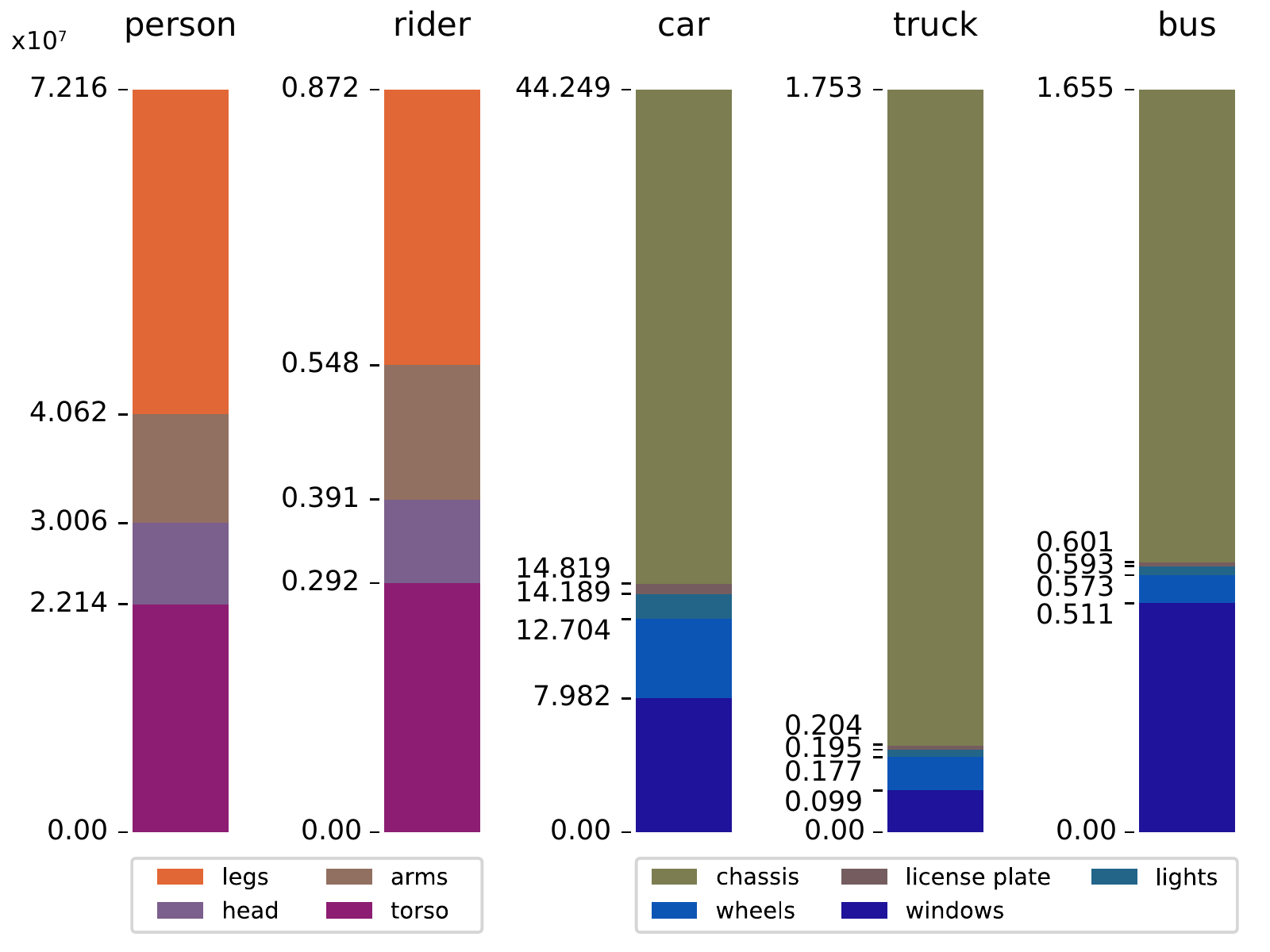}\\
	\end{center}
    \vspace{-15pt}
	\caption{Absolute number of Cityscapes pixels ($\times 10^7$) that we annotated per semantic class and per human/vehicle part.}
	\label{fig:num-pixels-cityscapes}
\end{figure}

Furthermore, for the five \textit{things} classes that we annotated on the parts level, we provide the absolute number of pixels per semantic class and per parts class in Figure~\ref{fig:num-pixels-cityscapes}. As can be seen, some part classes (bus lights, truck license plate) have very few pixels compared to others (vehicle chassis), and we anticipate interesting algorithms to be proposed to tackle this extreme imbalance.  

\section{PASCAL-Panoptic-Parts Dataset}
\label{sec:pascal-dataset}

\begin{figure}
	\centering
	\includegraphics[width=0.40\linewidth]{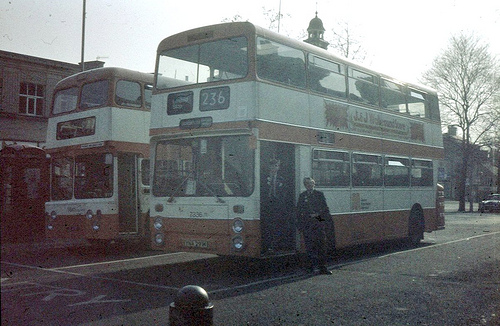}
	~\includegraphics[width=0.40\linewidth]{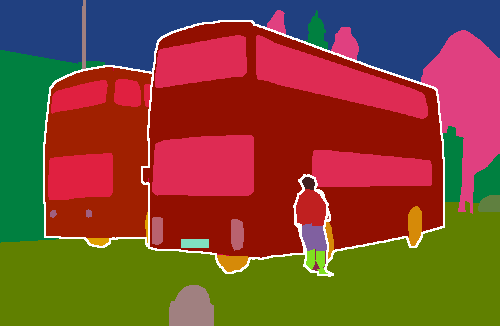}\\
	\includegraphics[width=0.40\linewidth]{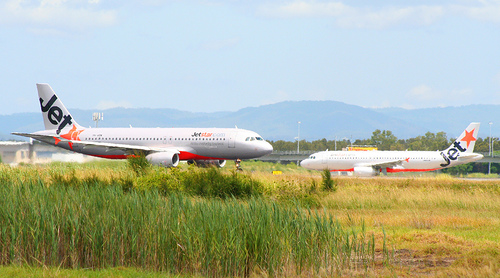}
	~\includegraphics[width=0.40\linewidth]{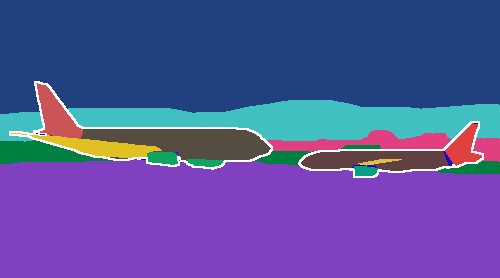}\\
	\includegraphics[width=0.40\linewidth]{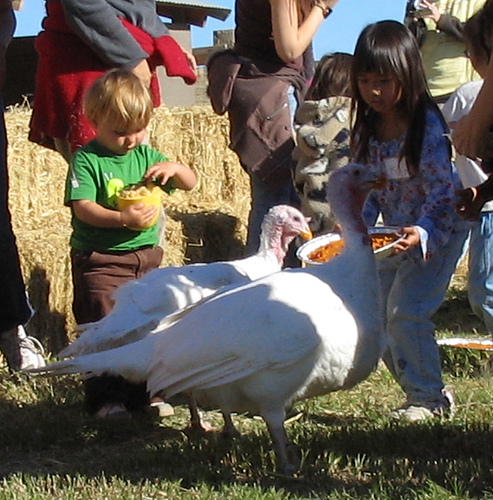}
	~\includegraphics[width=0.40\linewidth]{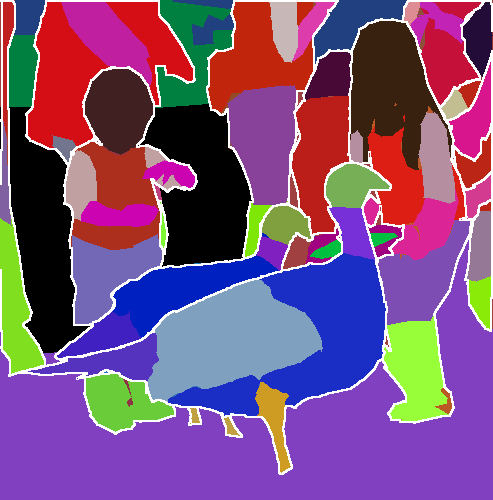}\\
	\includegraphics[width=0.40\linewidth]{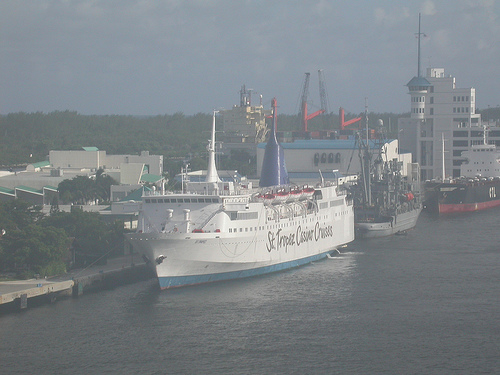}
	~\includegraphics[width=0.40\linewidth]{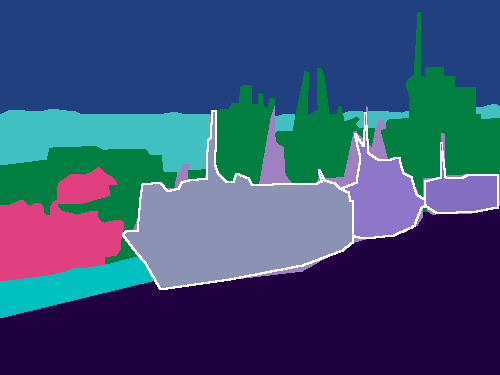}\\
	\includegraphics[width=0.40\linewidth]{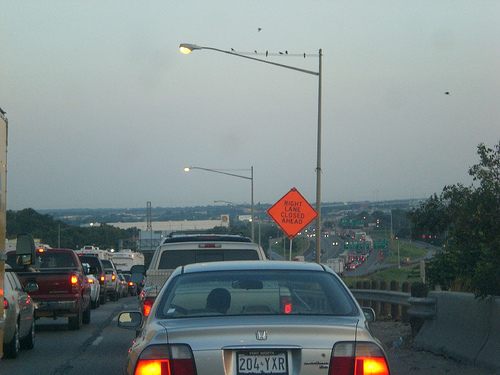}
	~\includegraphics[width=0.40\linewidth]{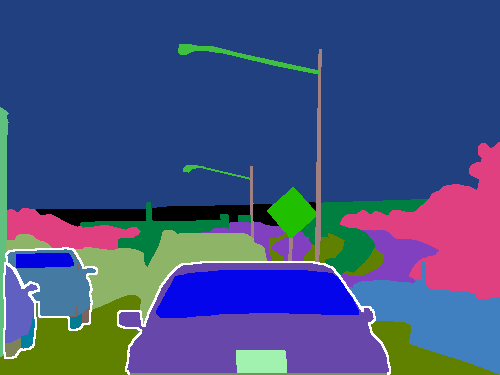}\\
	\vspace{-5pt}
	\caption{Examples of PASCAL-Panoptic-Parts images and labels from the training split. The benefits of our ``best-effort'' merging strategy are clear in the last two images, where the semantic-level labels (from \textit{PASCAL-Context}), \textit{boat} and \textit{car}, provide information for the unlabeled pixels of \textit{PASCAL-Parts}.}
	\label{fig:gt-examples-pascal}
\end{figure}

The annual PASCAL Visual Object Classes (VOC) challenges have been a major benchmark for visual object recognition and detection tasks, and its corresponding datasets serve as an essential resource for relevant researches in the field. The PASCAL VOC 2010 challenge \cite{pascal-voc-2010} has the main goal of recognizing several classes of visual objects in realistic scenes. It has 21738 images in total, which are split into three splits, i.e., training, validation, and test, containing 4998, 5105 and 11635 images, respectively.

We introduce \textit{PASCAL-Panoptic-Parts}, a summarized compound dataset that extends PASCAL VOC 2010 with panoptic-level annotations and part-level annotations, based on the extensions \textit{PASCAL-Context}~\cite{mottaghi14context} and \textit{PASCAL-Parts} \cite{chen2014detect}. In \textit{PASCAL-Panoptic-Parts} we include the training and validation splits (10103 images in total) since they are the only common subset between the three datasets. By combining the \textit{PASCAL-Context} and \textit{PASCAL-Parts} annotations, the resulting \textit{PASCAL-Panoptic-Parts} dataset has both \textit{stuff} and \textit{things} annotations, which is compatible with panoptic segmentation~\cite{Kirillov2019PS}. In Figure~\ref{fig:gt-examples-pascal}, we provide examples of images and labels from \textit{PASCAL-Panoptic-Parts}.

\textit{PASCAL-Panoptic-Parts} is labeled on semantic level with 80 \textit{stuff} classes (including one \textit{unlabeled/unknown} class) and 20 things classes. The 20 \textit{things} classes are also labeled on instance level, and parts level with 193 parts classes.
The total 100 semantic classes are a subset of the original \textit{PASCAL-Context}~\cite{mottaghi14context} 459 semantic classes, which are the top 100 classes selected based on the product of the average occupied area per image and the number of images it appears. The measure of the average occupied area of a class indicates the visual exposure of it in terms of pixels. It encodes the semantic importance of the class as it takes up a significant portion in the image it occurs. Selection based on the average occupied area ensures that the class distribution includes classes that are both generally frequent and obvious in the images and classes that are rare but are significantly obvious in the images. In this case, the class distribution is long-tailed, and the relatively rare classes can be included for diversity. \textit{PASCAL-Context} official set contains only 59 out of 459 classes, and we included all of them in our selection.

The 20 \textit{things} classes are common between \textit{PASCAL-Context} and \textit{PASCAL-Parts}. As it is difficult to assess the completeness and correctness of those annotations without a complete ground truth annotation set, we decided to keep labels from both datasets and count on a vigorous merging strategy in order to maximize the maintained label information. The benefits are visible in the last two images of Figure~\ref{fig:gt-examples-pascal}. The extent of some instance-level labeled regions is smaller than the corresponding object, leading the rest of the object to be labeled with semantic-level labels. If we had opted for ignoring the semantic-level labels for the \textit{things} classes of \textit{PASCAL-Context} we would have lost valuable information for a large number of pixels. According to our hierarchical labeling scheme we provide as much information as possible for the three levels of the hierarchy and we leave to the user of the dataset to select which level of information is needed or which some pixels should be ignored. 


\subsection{Merging process}
\label{ssec:merging}

\textit{PASCAL-Panoptic-Parts} is created by merging the \textit{PASCAL-Context} dataset with \textit{PASCAL-Parts}. \textit{PASCAL-Context} only contains semantic-level annotation for the semantic classes and \textit{PASCAL-Parts} only contains object-level annotations and/or part-level annotations for the 20 \textit{things} class (\textit{boat}, \textit{chair}, \textit{table} and \textit{sofa} classes only have object-level annotations).

During merging, we have three types of conflicts: firstly, because the two datasets have semantic classes in common, there are overlapping labels, which we define as inter-dataset conflicts. Secondly, given that \textit{PASCAL-Parts} 
is annotated object-wise and not image-wise, a pixel may have multiple \textit{instance ids}, which we define as object-level intra-dataset conflicts. The third type of overlap arises from partonomy in the part-level annotations from \textit{PASCAL-Parts} where some parts are part of others, i.e. \textit{nose} is a part of \textit{head}. In the part-level annotations from \textit{PASCAL-Parts}, since individual masks are annotated for each part class, having multiple hierarchical labels per pixels is allowed. However, in our definition, also seen in Section~\ref{sec: label format}, only one hierarchical label is allowed per pixel. These conflicts are defined as part-level intra-dataset conflicts.

We outline our merging strategy using an ``imaginary canvas'' $p$ for the pixel in question on which, in each step, we fuse information from the two datasets for each level of our label hierarchy. Furthermore, we denote the label from \textit{PASCAL-Context} as $PC$ and that from \textit{PASCAL-Parts} as $PP$.
We propose the following merging strategy to make the \textit{PASCAL-Panoptic-Parts} annotation compatible with the hierarchical panoptic format and to resolve the aforementioned conflicts in each step:


\renewcommand{\theenumii}{\roman{enumii}}
\begin{enumerate}

\item At the semantic level, we first select 100 semantic classes (informative labels) out of 459 from \textit{PASCAL-Context} based on the selection criterion and treat the other classes as \textit{unlabeled/unknown}. Then for each pixel in an image, we have the following cases:
\begin{enumerate}
    \item Both $PC$ and $PP$ are informative (not \textit{unknown/unlabeled}), and the labels agree, then we fill $p$ with this label; if the labels do not agree, we use $PP$ with higher priority. This is because \textit{PASCAL-Parts} is essentially encoding more in-depth information (object-level and/or part-level) compared to \textit{PASCAL-Context}, which is what we aim prefer to include. By using the semantic label of $PP$, we ensure that in the following merging steps we can include object-level annotations and potentially part-level annotations as much as possible.
    \item $PC$ is informative but $PP$ is not, then we fill $p$ with $PC$;
    \item $PP$ is informative but $PC$ is not, then we fill $p$ with $PP$.
\end{enumerate}
which resolves the inter-dataset conflicts.


\item At the instance level, because \textit{PASCAL-Parts} only provides object-level annotations, we first instantiate the semantic classes by grouping the object-level annotations by their semantic classes and yield the instance indices. Subsequently, we extend the semantic class labels to instance-level by appending the instance indices. With regarding this extended instance-level annotation as the flattened \textit{PASCAL-Parts} annotation $PP_{flat}$, for each pixel in the image, we iterate through the object-level annotations from \textit{PASCAL-Parts} by the provided object order, and if $p$ is annotated with $PP$ in the previous step, we fill it with $PP_{flat}.$
During this step, as we iterate through the order of objects provided in the \textit{PASCAL-Parts}, if a pixel originally belongs to multiple objects, it will be assigned with the label of the last object following this order. This process then resolves the object-level intra-dataset conflicts.

\item At the part level, for each pixel in the image, if $p$ is annotated with $PP_{flat}$ from the previous step and its corresponding semantic class belongs to one of the \textit{things} classes that contains part-level information, we first extend the annotation by appending the part-level \textit{unlabeled/unknown} class labels. Then we annotate $p$ with specifying the priority of the part-level classes according to partonomy, i.e. we impose the part-level class \textit{nose} over \textit{head}, in order to resolve the part-level intra-dataset conflicts.

\end{enumerate}

\begin{figure}[t]
	\centering
	\includegraphics[width=1.0\linewidth, trim={0.cm 0.2cm 0.0cm 0.0cm}, clip]{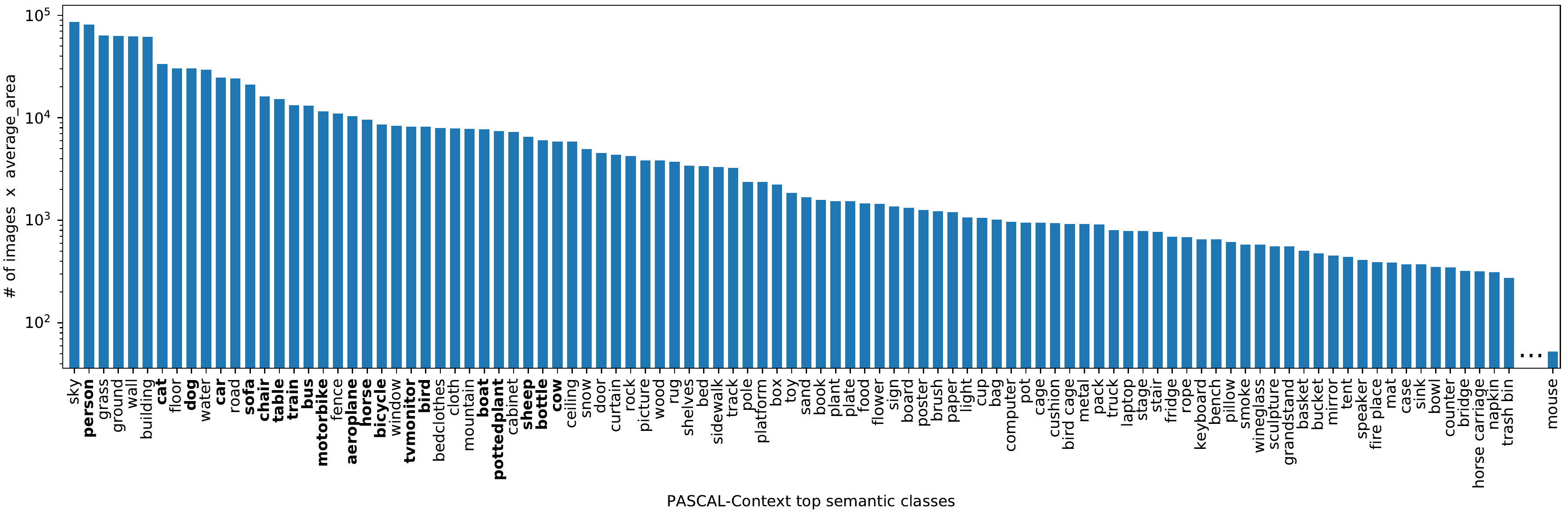}
	\vspace{-15px}
	\caption{Selection criterion for choosing the 100 semantic classes to include in our dataset. The classes in \textbf{bold} denote \textit{things} classes and have instances as well as part-level labels.}
	\label{fig:classes-selection}
\end{figure}



\subsection{Dataset Statistics}
\begin{figure}[t]
	\centering
	\includegraphics[width=1.0\linewidth, trim={0.cm 0.2cm 0.0cm 0.0cm}, clip]{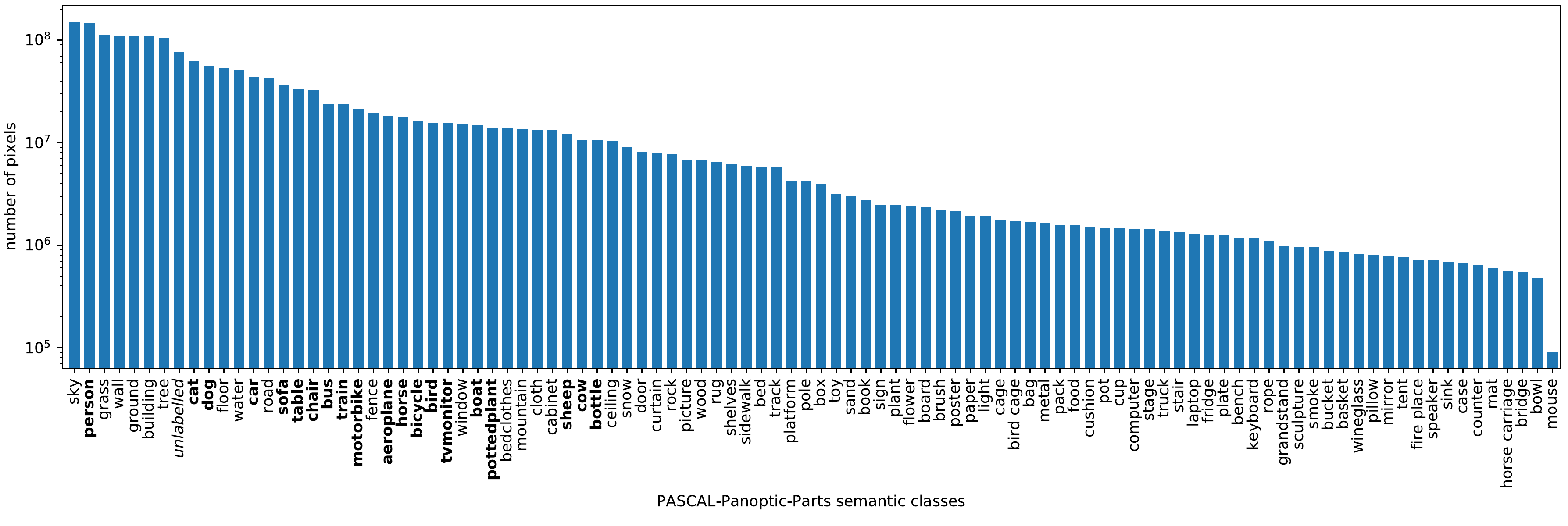}
	\vspace{-15px}
	\caption{Number of pixels for PASCAL-Panoptic-Parts 100 semantic class labels for training and validation splits (10103 images). The classes in \textbf{bold} denote \textit{things} classes and have instances as well as part-level labels.}
	\label{fig:classes-pascal}
\end{figure}

\begin{figure}[t]
	\centering
	\includegraphics[width=0.65\linewidth, trim={1.5cm 1.0cm 1.0cm 1.0cm}, clip]{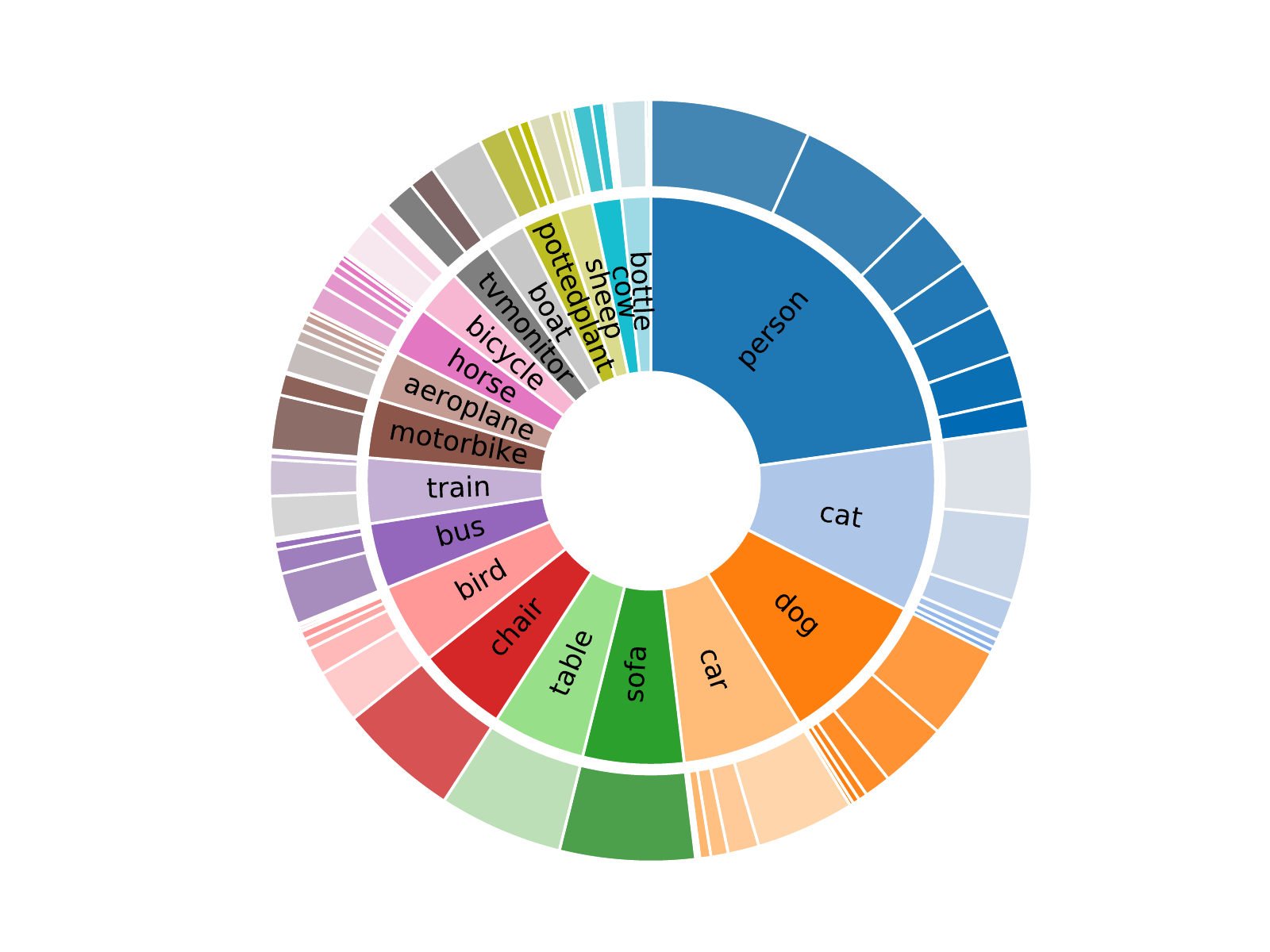}
	\vspace{-5px}
	\caption{The 20 \textit{things} classes (out of total 100 semantic classes) of PASCAL-Panoptic-Parts in the inner circle and their respective parts classes in the outer circle order by the number of pixels in the dataset. For some classes~e.g., person, the parts classes are grouped for cleaner visualization. Colors do not to the colormap used for visualizations.}
	\label{fig:things-classes-pascal}
\end{figure}

In Figure~\ref{fig:classes-selection}, the selection criterion is plotted against all the 99 semantic classes. The top 98 semantic classes (+1 unlabeled/unknown -- not shown in Figure) are selected together with the \textit{mouse} class that scores much lower (\#166), but was included, so all classes from the official set of 59 classes are included. In Figure~\ref{fig:classes-pascal}, the absolute number of pixels, is shown for all 100 semantic classes for the 10103 images of the training and validation splits. Figures~\ref{fig:classes-selection} and~\ref{fig:classes-pascal} differ in the sorting order of classes due to the merging strategy explained in Section~\ref{ssec:merging}. Finally, in Figure~\ref{fig:things-classes-pascal} the 20 \textit{things} classes that are labeled instance-wise and 16 of them part-wise are shown, together with a summarization of their 193 part-level classes to 73 for easier usage visualization. The classes of the reduced set can be found at \href{https://github.com/tue-mps/panoptic_parts}{github.com/tue-mps/panoptic\_parts}.

\subsection{Future improvements}
In addition, further exploration in the measure for selecting feasible semantic classes from \textit{PASCAL-Context} can be done to include more semantically important classes, i.e., taking both the number of occurrences and the average occupied area into consideration. A refined version of the \textit{PASCAL-Panoptic-Parts} dataset will also be investigated in future work.

\section{Hierarchical label format} \label{sec: label format}
We decided to extend the Cityscapes dataset~\cite{Cordts2016Cityscapes} label format due to its compactness and directness and include part-level labels in a hierarchical manner. Cityscapes dataset is labeled pixel-wise with an integer (base 10) \textit{id}, which has up to 5 digits. Every pixel in an image has a \textit{semantic id} (0-99), encoding either \textit{things} or \textit{stuff} semantic classes, e.g., car, person, building, traffic light. If a pixel belongs to a countable object (\textit{thing}), it may also have an \textit{instance id} (0-999), encoding different instances of the same semantic class in an image. The semantic classes are a fixed, predefined set for the whole dataset. The \textit{instance id} is a counter per \textit{things} semantic class and per image.

We decided to extend this format with a two-digit \textit{part id} (0-99) denoting the semantic part-level class labels. This format enables to define up to 100 parts classes for every \textit{things} semantic class. Moreover, the parts are bounded to a specific instance, which makes our format compatible with the recently introduced instance-wise object parsing task~\cite{li2017holistic, zhao2018understanding, gong2018instance}.

To summarize, each pixel in our hierarchical label format has:
\begin{itemize}[noitemsep, topsep=0pt]
	\item An up to 2-digit \textit{semantic id}, encoding a \textit{things} or \textit{stuff} semantic class.
\end{itemize}

If the pixel belongs to a \textit{things} semantic class it can optionally have:
\begin{itemize}[noitemsep, topsep=0pt]
	\item An up to 3-digit \textit{instance id}, a counter of instances per image.
\end{itemize}

Finally, if the pixel belongs to a \textit{things} semantic class and is labeled instance-wise it can optionally have:
\begin{itemize}[noitemsep, topsep=0pt]
	\item An up to 2-digit \textit{part id}, encoding the parts semantic class per-instance and per-image.
\end{itemize}

We compactly encode the aforementioned \textit{ids} into an up to 7-digit \textit{id}, for which the first two digits (starting from the left) encode the semantic class, the next 3 encode the instance (after zero pre-padding), and the final two encode the parts class (after zero pre-padding). We use the following formula, which produces \textit{ids} that can be stored in a single image-like file:

\noindent
{\small
$
\textit{id} = 
\begin{cases}
(\textit{semantic id}) & \text{{\footnotesize semantic level}} \\ 
(\textit{semantic id}) \cdot 10^3 + (\textit{instance id}) & \text{{\footnotesize semantic, instance levels}} \\ 
(\textit{semantic id}) \cdot 10^5 + (\textit{instance id}) \cdot 10^2 + (\textit{part id}) & \text{{\footnotesize semantic, instance, parts levels}}
\end{cases}
$
}

For example, for \textit{Cityscapes-Panoptic-Parts} a sky (\textit{stuff}) pixel will have $id = 23$, a car (\textit{things}) pixel that is labeled only on the semantic level will have $id = 26$, if it's labeled also on instance level it can have $id = 26002$, and a person (\textit{things}) pixel that is labeled on all three levels (of Figure~\ref{fig:tasks}) can have $id = 2401002$.

We handle the unlabeled/void/``do not care pixels'' in the three levels as follows:
\begin{itemize}
\item Semantic level: For \textit{Cityscapes-Panoptic-Parts} we use the original Cityscapes void class. For \textit{PASCAL-Panoptic-Parts} we use the class with $id = 0$ (first class).
\item Instance level: For instances the void class is not needed. If a pixel does not belong to an object or cannot be labeled on instance level then it has only an up to 2-digit \textit{semantic id}.
\item Parts level: For both datasets we use the convention that, for each semantic class, the part-level class with $id = 0$ (first class) represents the void pixels~e.g. for a person pixel, $id = 2401000$ represents the void parts pixels of instance $10$. The need for a void class arises during the manual annotation process but in principle it is not needed at the parts level. Thus, we try to minimize void parts level pixels and assign them instead only the semantic and/or instance level labels.
\end{itemize}

\section{Acknowledgments}
We kindly thank Chenyang Lu, Xiaoxiao Wen, and Gijs Kok, Guus Engels, for providing a large number of initial annotations. Moreover, we thank the following students for annotating the rest of the dataset: 
Iko Vloothuis, 
Quinten Aalders, 
Gijs Cunnen, 
Noud van de Gevel, 
Jelin Leslin, 
Daan Hommersom, 
and 4 more who wished to maintain anonymity.

\section{More examples}
\begin{figure}
	\centering
	\includegraphics[width=0.48\linewidth]{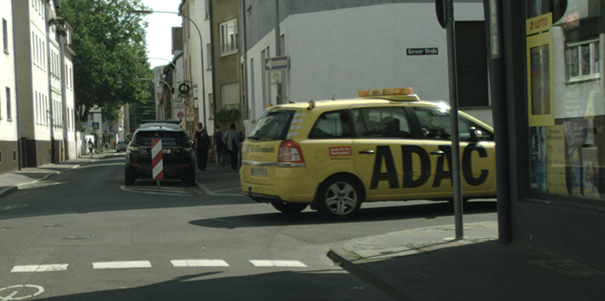}
	\includegraphics[width=0.48\linewidth]{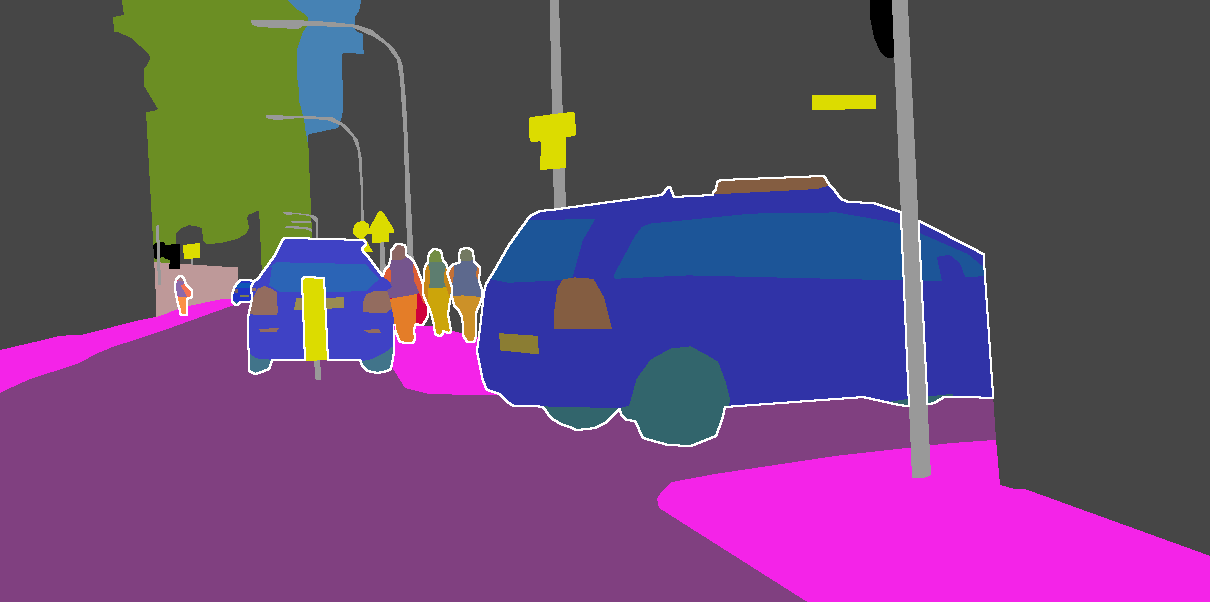}\\ 
	\includegraphics[width=0.48\linewidth]{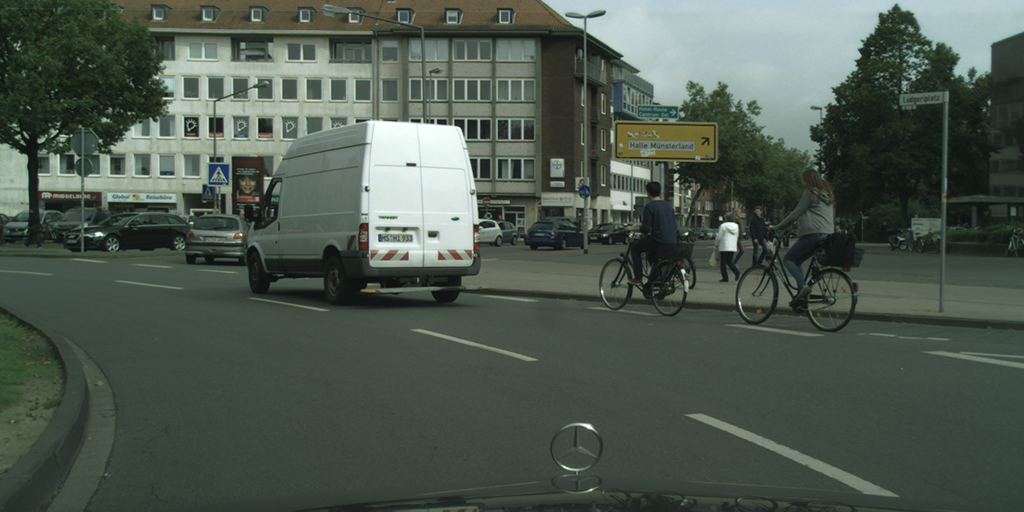}
	\includegraphics[width=0.48\linewidth]{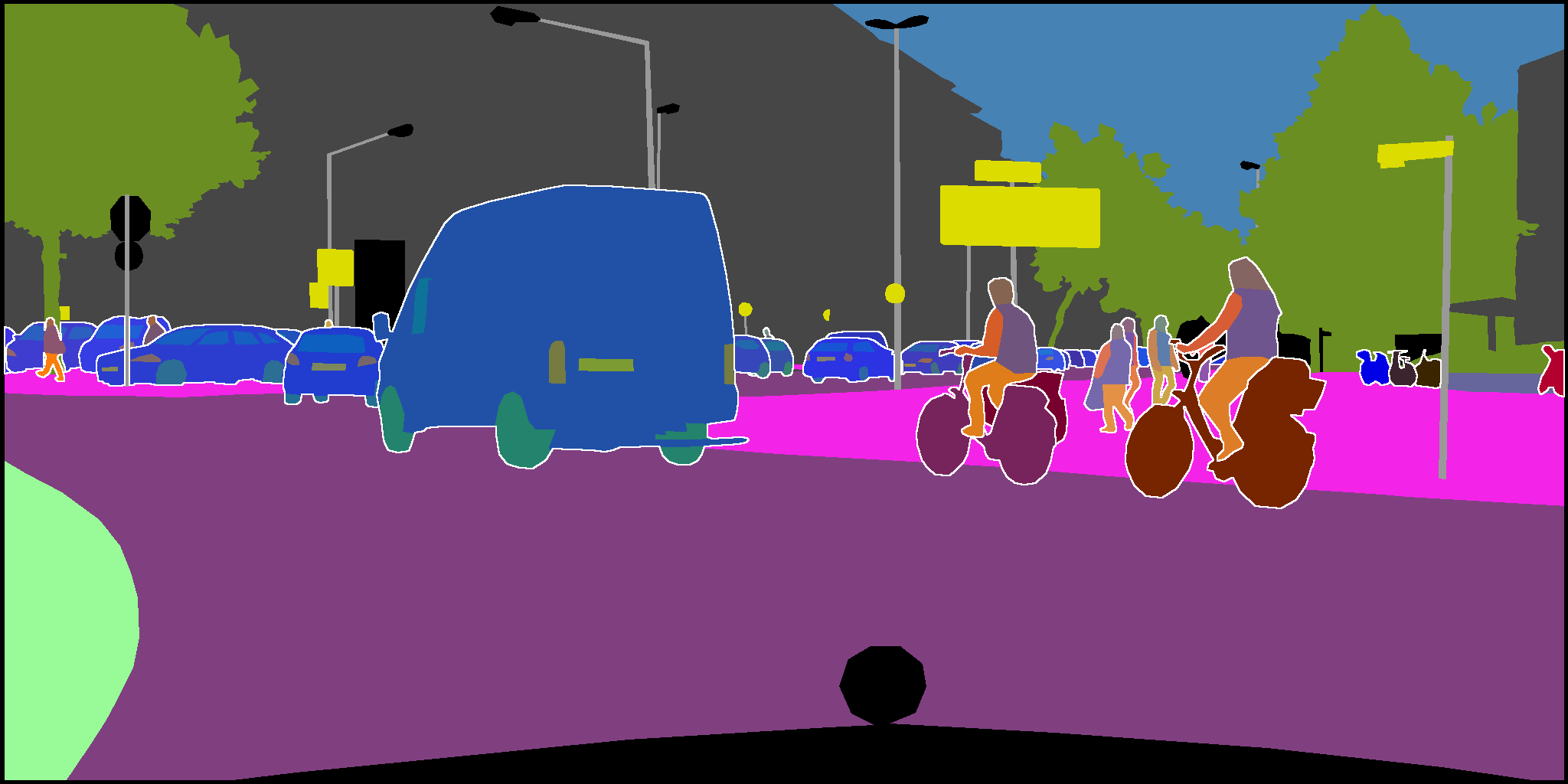}\\ 
	\includegraphics[width=0.48\linewidth]{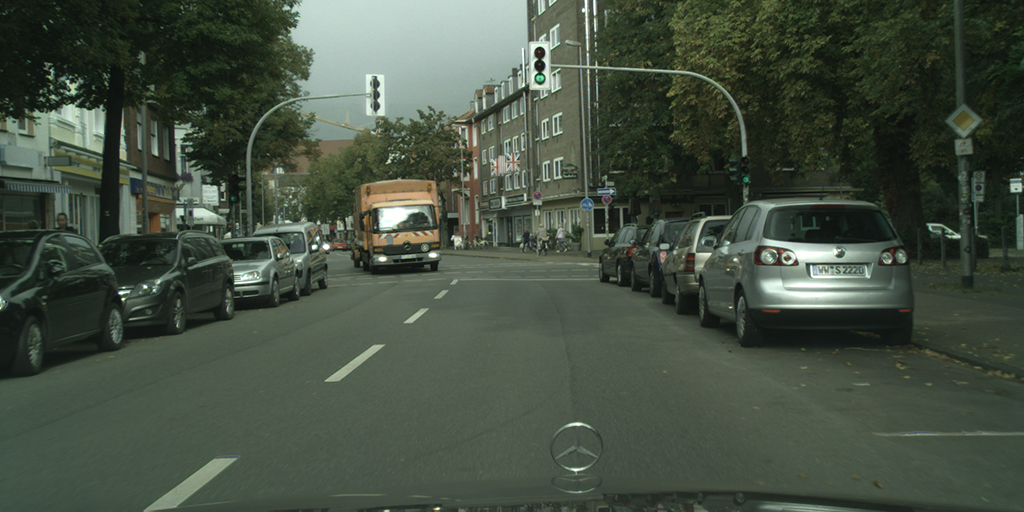}
	\includegraphics[width=0.48\linewidth]{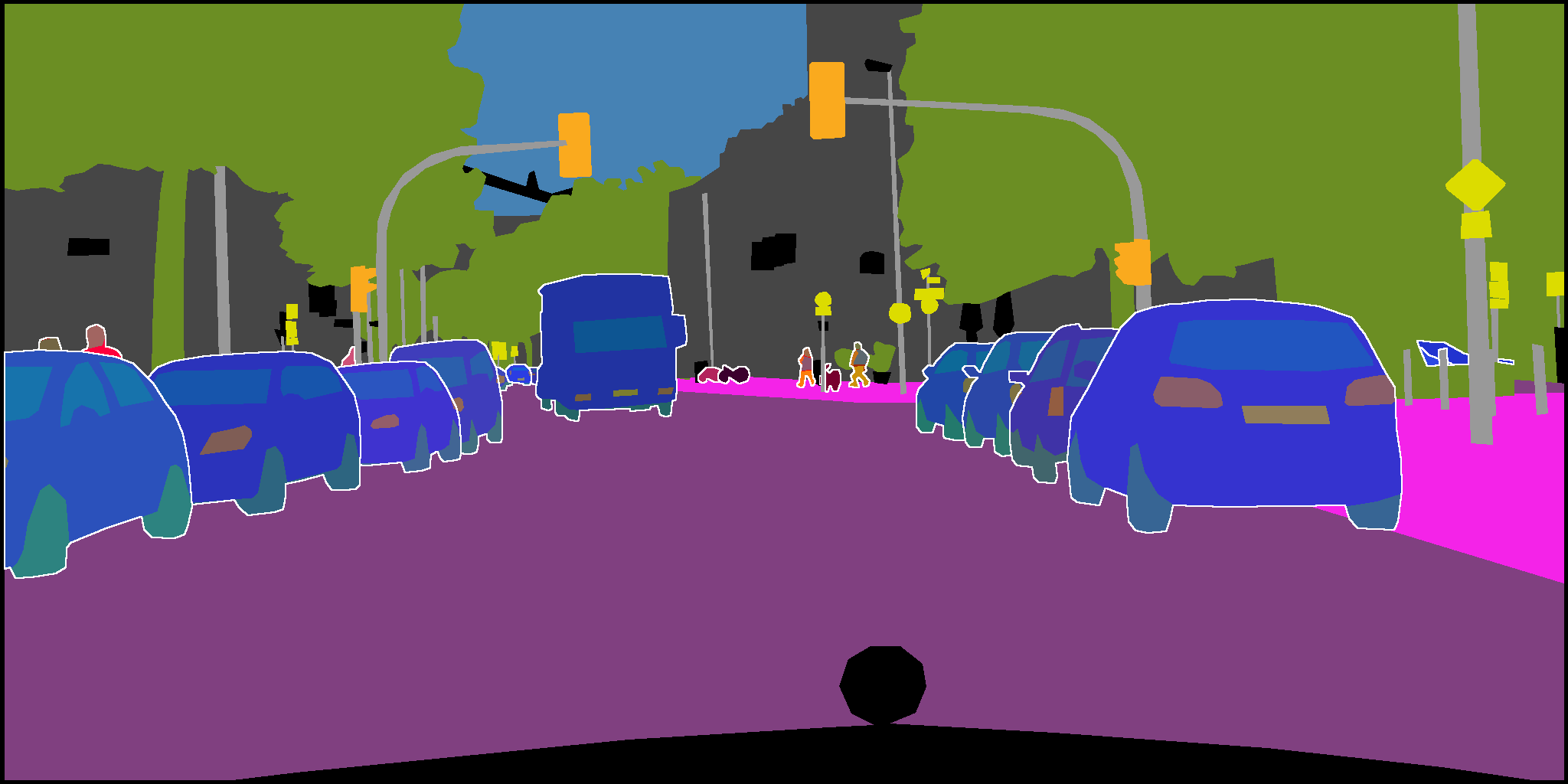}\\ 
	\includegraphics[width=0.48\linewidth]{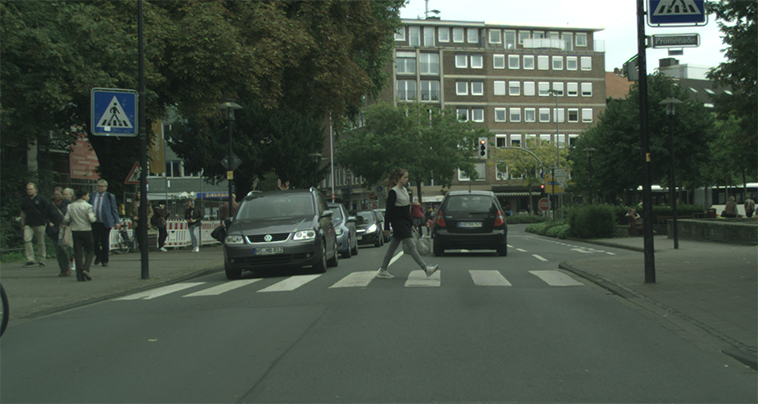}
	\includegraphics[width=0.48\linewidth]{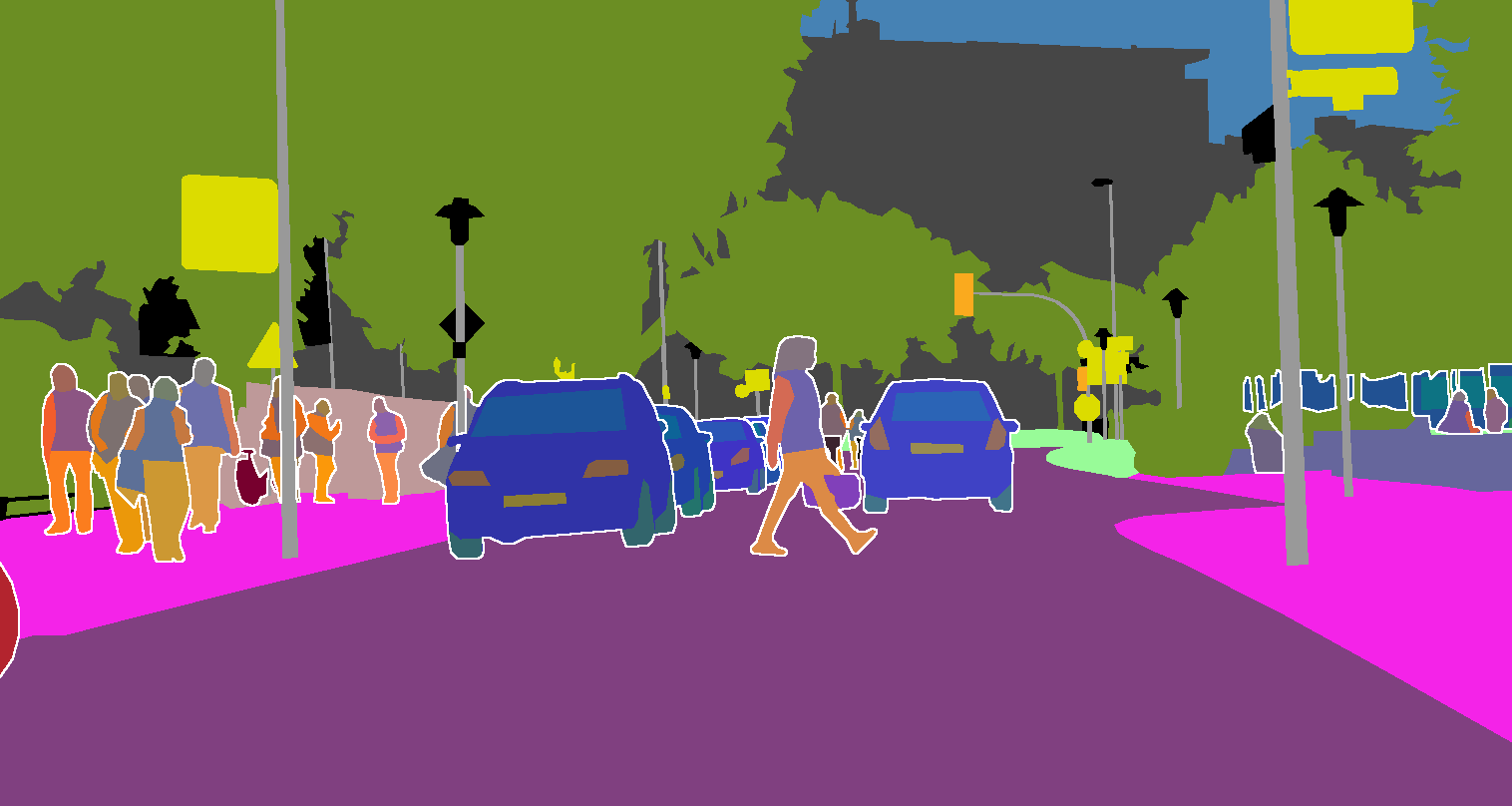}\\ 
	\includegraphics[width=0.48\linewidth]{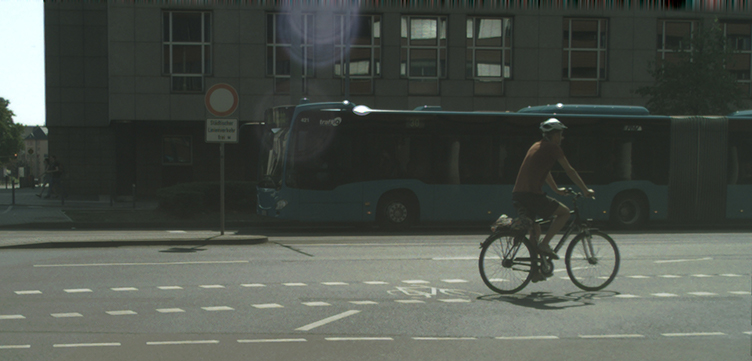}
	\includegraphics[width=0.48\linewidth]{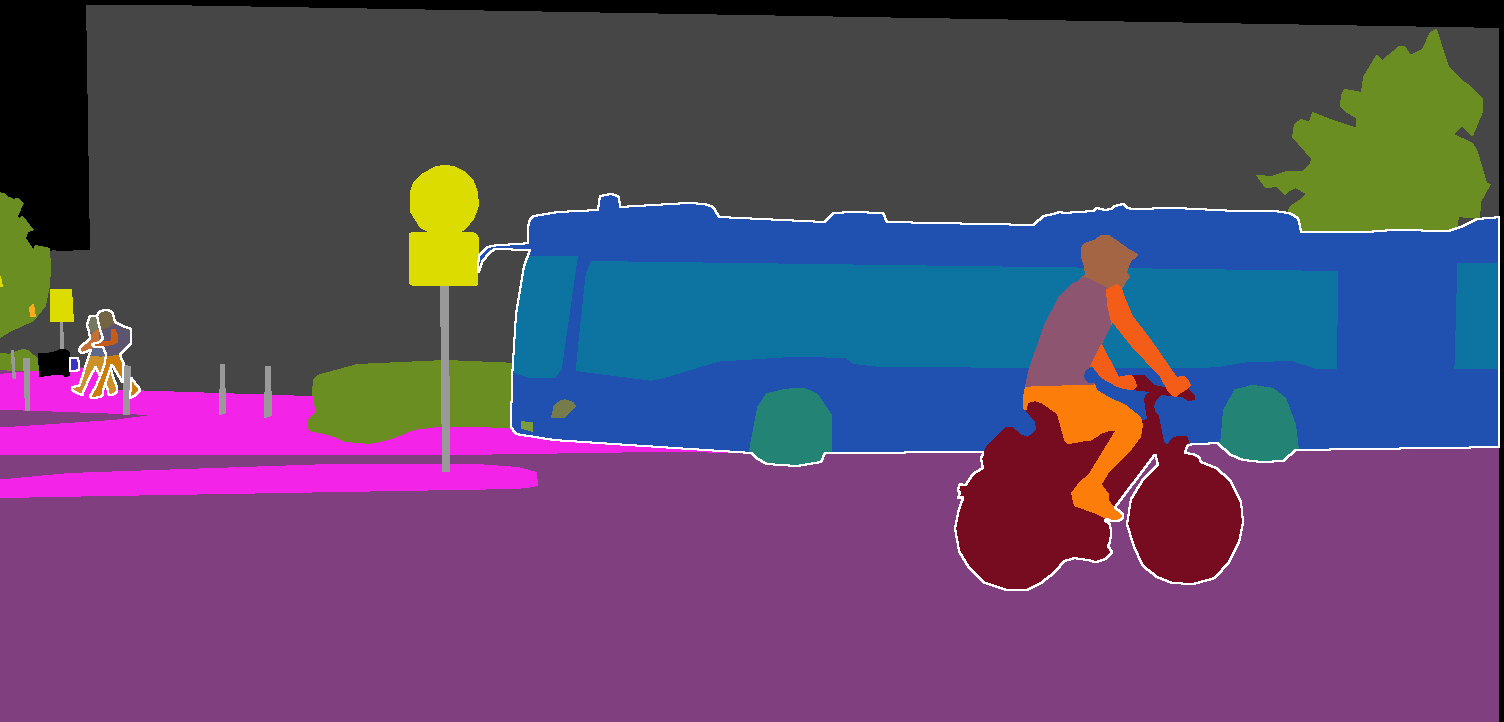}\\
	\includegraphics[width=0.48\linewidth]{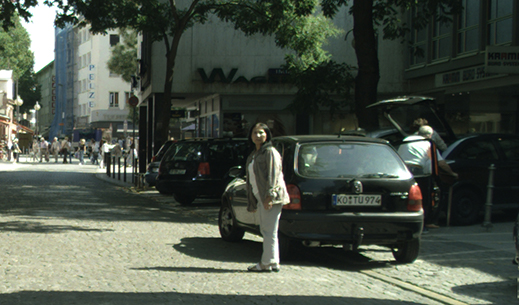}
	\includegraphics[width=0.48\linewidth]{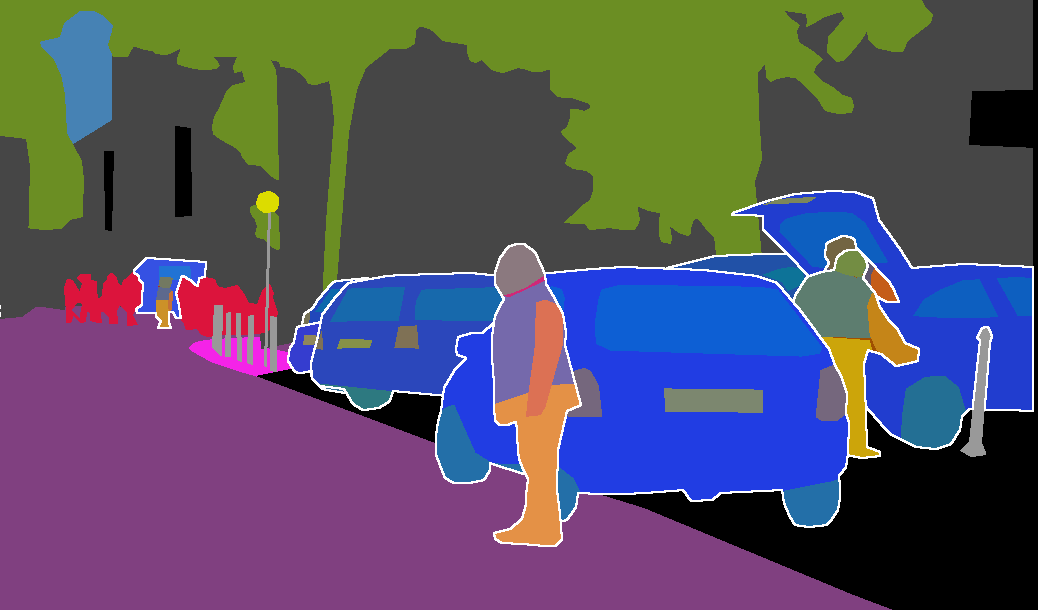}\\
	\caption{More examples from \textit{Cityscapes-Panoptic-Parts}.}
	\label{fig:more_examples}
\end{figure}

\begin{figure}
	\centering
	\includegraphics[width=0.35\linewidth]{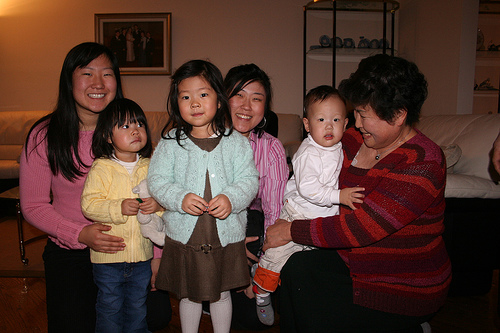}
	\includegraphics[width=0.35\linewidth]{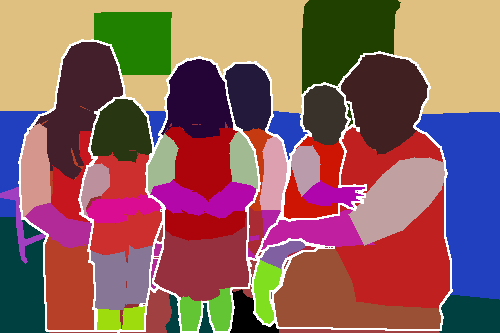}\\
	\includegraphics[width=0.35\linewidth]{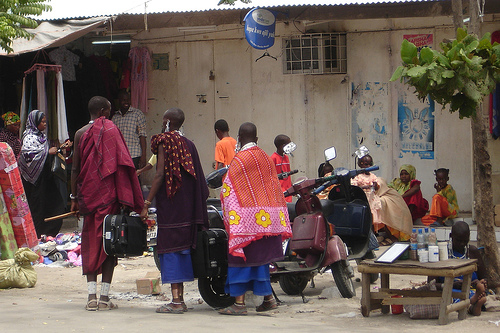}
	\includegraphics[width=0.35\linewidth]{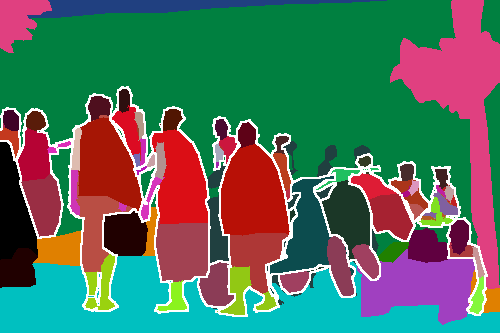}\\
	\includegraphics[width=0.35\linewidth]{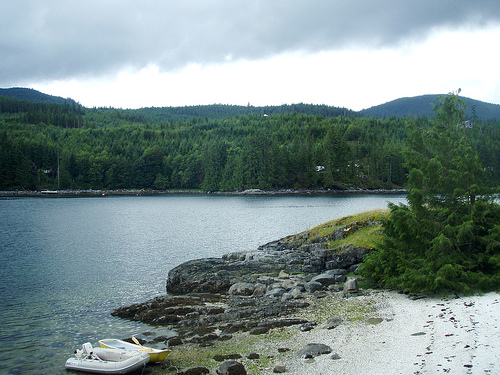}
	\includegraphics[width=0.35\linewidth]{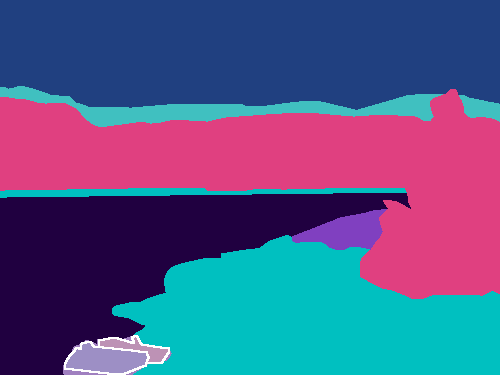}\\
	\includegraphics[width=0.35\linewidth]{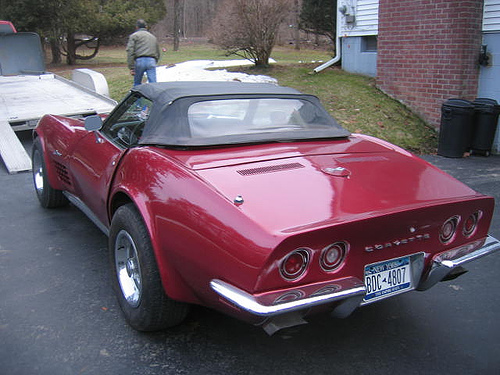}
	\includegraphics[width=0.35\linewidth]{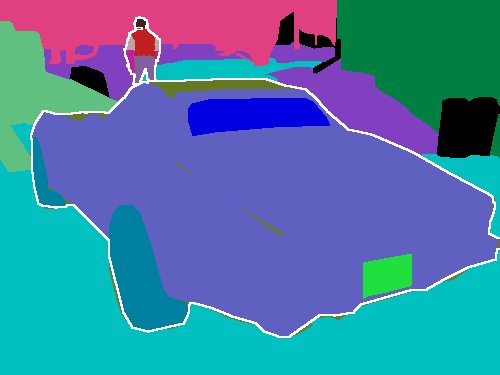}\\
	\includegraphics[width=0.35\linewidth]{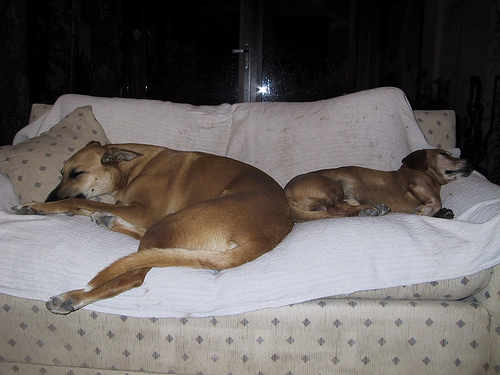}
	\includegraphics[width=0.35\linewidth]{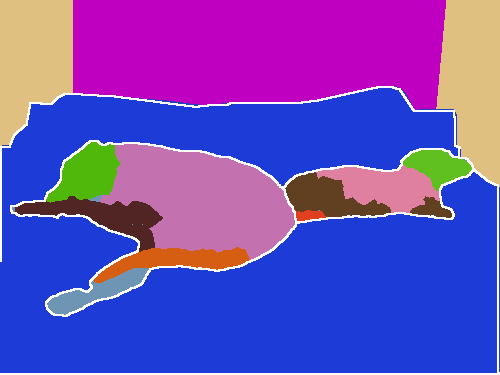}\\
	\includegraphics[width=0.35\linewidth]{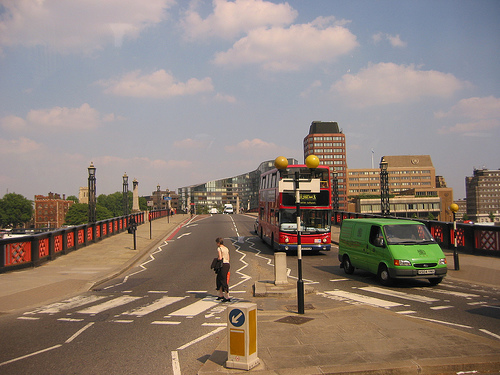}
	\includegraphics[width=0.35\linewidth]{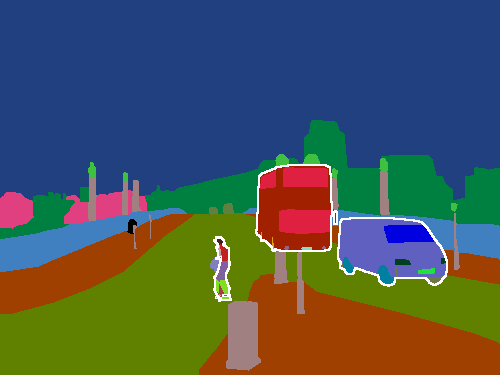}\\
	\caption{More examples from \textit{PASCAL-Panoptic-Parts}.}
	\label{fig:more-examples-pascal}
\end{figure}

In Figures~\ref{fig:more_examples} and~\ref{fig:more-examples-pascal}, more examples of images and corresponding labels from the two datasets are shown.

{\small
	\bibliographystyle{ieee_fullname}
	\bibliography{biblio}
}

\pagebreak

\appendix

\section{Semantic classes and part-level classes names}
In this Section we list all the \textit{things} semantic classes that have part-level class labels in the two datasets. The format is the following:

\dirtree{%
.1 Dataset.
  .2 semantic level class.
    .3 parts level class.
}

\subsection{Cityscapes-Panoptic-Parts}
{\footnotesize
\dirtree{%
.1 Cityscapes-Panoptic-Parts \textit{things} classes with part-level classes.
  .2 person.
    .3 torso.
    .3 head.
    .3 arm.
    .3 leg.
  .2 rider.
    .3 torso.
    .3 head.
    .3 arm.
    .3 leg.
  .2 car.
    .3 window.
    .3 wheel.
    .3 light.
    .3 license plate.
    .3 chassis.
  .2 truck.
    .3 window.
    .3 wheel.
    .3 light.
    .3 license plate.
    .3 chassis.
  .2 bus.
    .3 window.
    .3 wheel.
    .3 light.
    .3 license plate.
    .3 chassis.
}
}

\subsection{PASCAL-Panoptic-Parts}
{\footnotesize
\dirtree{%
.1 PASCAL-Panoptic-Parts \textit{things} classes.
  .2 aeroplane.
    .3 body.
    .3 stern.
    .3 lwing.
    .3 rwing.
    .3 tail.
    .3 engine.
    .3 wheel.
  .2 bicycle.
    .3 fwheel.
    .3 bwheel.
    .3 saddle.
    .3 handlebar.
    .3 chainwheel.
    .3 headlight.
  .2 bird.
    .3 head.
    .3 leye.
    .3 reye.
    .3 beak.
    .3 torso.
    .3 neck.
    .3 lwing.
    .3 rwing.
    .3 lleg.
    .3 lfoot.
    .3 rleg.
    .3 rfoot.
    .3 tail.
  .2 boat.
  .2 bottle.
    .3 cap.
    .3 body.
  .2 bus.
    .3 frontside.
    .3 leftside.
    .3 rightside.
    .3 backside.
    .3 roofside.
    .3 leftmirror.
    .3 rightmirror.
    .3 fliplate.
    .3 bliplate.
    .3 door.
    .3 wheel.
    .3 headlight.
    .3 window.
  .2 car.
    .3 frontside.
    .3 leftside.
    .3 rightside.
    .3 backside.
    .3 roofside.
    .3 leftmirror.
    .3 rightmirror.
    .3 fliplate.
    .3 bliplate.
    .3 door.
    .3 wheel.
    .3 headlight.
    .3 window.
  .2 cat.
    .3 head.
    .3 leye.
    .3 reye.
    .3 lear.
    .3 rear.
    .3 nose.
    .3 torso.
    .3 neck.
    .3 lfleg.
    .3 lfpa.
    .3 rfleg.
    .3 rfpa.
    .3 lbleg.
    .3 lbpa.
    .3 rbleg.
    .3 rbpa.
    .3 tail.
  .2 chair.
  .2 cow.
    .3 head.
    .3 leye.
    .3 reye.
    .3 lear.
    .3 rear.
    .3 muzzle.
    .3 lhorn.
    .3 rhorn.
    .3 torso.
    .3 neck.
    .3 lfuleg.
    .3 lflleg.
    .3 rfuleg.
    .3 rflleg.
    .3 lbuleg.
    .3 lblleg.
    .3 rbuleg.
    .3 rblleg.
    .3 tail.
  .2 table.
  .2 dog.
    .3 head.
    .3 leye.
    .3 reye.
    .3 lear.
    .3 rear.
    .3 nose.
    .3 torso.
    .3 neck.
    .3 lfleg.
    .3 lfpa.
    .3 rfleg.
    .3 rfpa.
    .3 lbleg.
    .3 lbpa.
    .3 rbleg.
    .3 rbpa.
    .3 tail.
    .3 muzzle.  
  .2 horse.
    .3 head.
    .3 leye.
    .3 reye.
    .3 lear.
    .3 rear.
    .3 muzzle.
    .3 lfho.
    .3 rfho.
    .3 lbho.
    .3 rbho.
    .3 torso.
    .3 neck.
    .3 lfuleg.
    .3 lflleg.
    .3 rfuleg.
    .3 rflleg.
    .3 lbuleg.
    .3 lblleg.
    .3 rbuleg.
    .3 rblleg.
    .3 tail.    
  .2 motorbike.
    .3 fwheel.
    .3 bwheel.
    .3 handlebar.
    .3 saddle.
    .3 headlight.    
  .2 person.
    .3 head.
    .3 leye.
    .3 reye.
    .3 lear.
    .3 rear.
    .3 lebrow.
    .3 rebrow.
    .3 nose.
    .3 mouth.
    .3 hair.
    .3 torso.
    .3 neck.
    .3 llarm.
    .3 luarm.
    .3 lhand.
    .3 rlarm.
    .3 ruarm.
    .3 rhand.
    .3 llleg.
    .3 luleg.
    .3 lfoot.
    .3 rlleg.
    .3 ruleg.
    .3 rfoot.  
  .2 pottedplant.
    .3 pot.
    .3 plant.    
  .2 sheep.
    .3 head.
    .3 leye.
    .3 reye.
    .3 lear.
    .3 rear.
    .3 muzzle.
    .3 lhorn.
    .3 rhorn.
    .3 torso.
    .3 neck.
    .3 lfuleg.
    .3 lflleg.
    .3 rfuleg.
    .3 rflleg.
    .3 lbuleg.
    .3 lblleg.
    .3 rbuleg.
    .3 rblleg.
    .3 tail.  
  .2 sofa.
  .2 train.
    .3 head.
    .3 hfrontside.
    .3 hleftside.
    .3 hrightside.
    .3 hbackside.
    .3 hroofside.
    .3 headlight.
    .3 coach.
    .3 cfrontside.
    .3 cleftside.
    .3 crightside.
    .3 cbackside.
    .3 croofside.
  .2 tvmonitor.
    .3 screen.
}
}

\end{document}